\pdfoutput=1

\documentclass[11pt]{article}

\PassOptionsToPackage{table}{xcolor}
\usepackage{acl}
\usepackage{times}
\usepackage{float}
\usepackage{latexsym}
\usepackage{amsmath}
\usepackage{amsfonts}
\usepackage{graphicx} 
\usepackage{longtable}
\usepackage{bbm}
\usepackage{multirow}
\usepackage{makecell}
\usepackage{algorithm}
\usepackage{algpseudocode}
\usepackage{array}
\usepackage{amssymb}
\usepackage{pifont}
\usepackage{booktabs}
\usepackage{tabularx}
\usepackage{enumitem} 
\definecolor{topone}{RGB}{255,235,235}
\definecolor{toptwo}{RGB}{235,245,255}

\newcolumntype{Y}{>{\raggedright\arraybackslash}X}

\usepackage[T1]{fontenc}

\usepackage[utf8]{inputenc}

\usepackage{microtype}

\usepackage{inconsolata}

\title{Towards Better Legal Reasoning LLMs: Signal Balancing and Reward Scheduling}

\author{
\textbf{Yinghao Hu}\textsuperscript{\textbf{1}},
\textbf{Qian Xiao}\textsuperscript{\textbf{2}},
\textbf{Yuting Huang}\textsuperscript{\textbf{1}},
\textbf{Kaisong Song}\textsuperscript{\textbf{2,4}},
\textbf{Yiquan Wu}\textsuperscript{\textbf{1,*}}
\\
\textbf{Ang Li}\textsuperscript{\textbf{3}},
\textbf{Junjie Cao}\textsuperscript{\textbf{2}},
\textbf{Jun Lin}\textsuperscript{\textbf{2}},
\textbf{Kun Kuang}\textsuperscript{\textbf{1,*}},
\textbf{Fei Wu}\textsuperscript{\textbf{1}}
\\[0.45em]
\textsuperscript{1}Zhejiang University, Hangzhou, China
\\
\textsuperscript{2}Alibaba Group, Hangzhou, China
\\
\textsuperscript{3}University of Macau, Macao, China
\\
\textsuperscript{4}Northeastern University, Shenyang, China
\\[0.35em]
\textbf{Emails:}
\texttt{\{huyinghao,wuyiquan,kunkuang\}@zju.edu.cn}
}

\begin{document}

\maketitle
\renewcommand{\thefootnote}{\fnsymbol{footnote}}
\footnotetext[1]{Corresponding authors.}
\renewcommand{\thefootnote}{\arabic{footnote}}

\begin{abstract}

In the legal domain, LLMs can improve reliability and interpretability by generating explicit reasoning traces. However, training LLMs to produce high-quality reasoning traces remains challenging. Existing supervised fine-tuning methods struggle when reasoning traces are significantly longer than final answers, as the learning signal for the answer becomes diluted. Meanwhile, reinforcement learning methods such as Group Relative Policy Optimization also have several limitations, including costly reward design, performance plateaus, and reward hacking. We therefore propose a two-stage training framework. In Stage I, JurisCoT-SFT employs a length-normalized dual objective to balance learning signals between reasoning traces and final answers. In Stage II, Lifecycle-Aware Backtrackable Policy Optimization dynamically evaluates and retires auxiliary reward signals based on their impact on final evaluation objectives, enabling efficient reward utilization. Our framework substantially improves legal reasoning performance on relevant tasks selected from multiple legal benchmarks, including ChJGBench, LawBench, and LegalBench. Experimental results demonstrate consistent improvements in both reasoning trace quality and final answer accuracy, while our models remain competitive with larger models.

\end{abstract}

\section{Introduction}

\begin{figure}[t]
    \centering
    \includegraphics[width=\columnwidth, trim=5 5 5 5, clip]{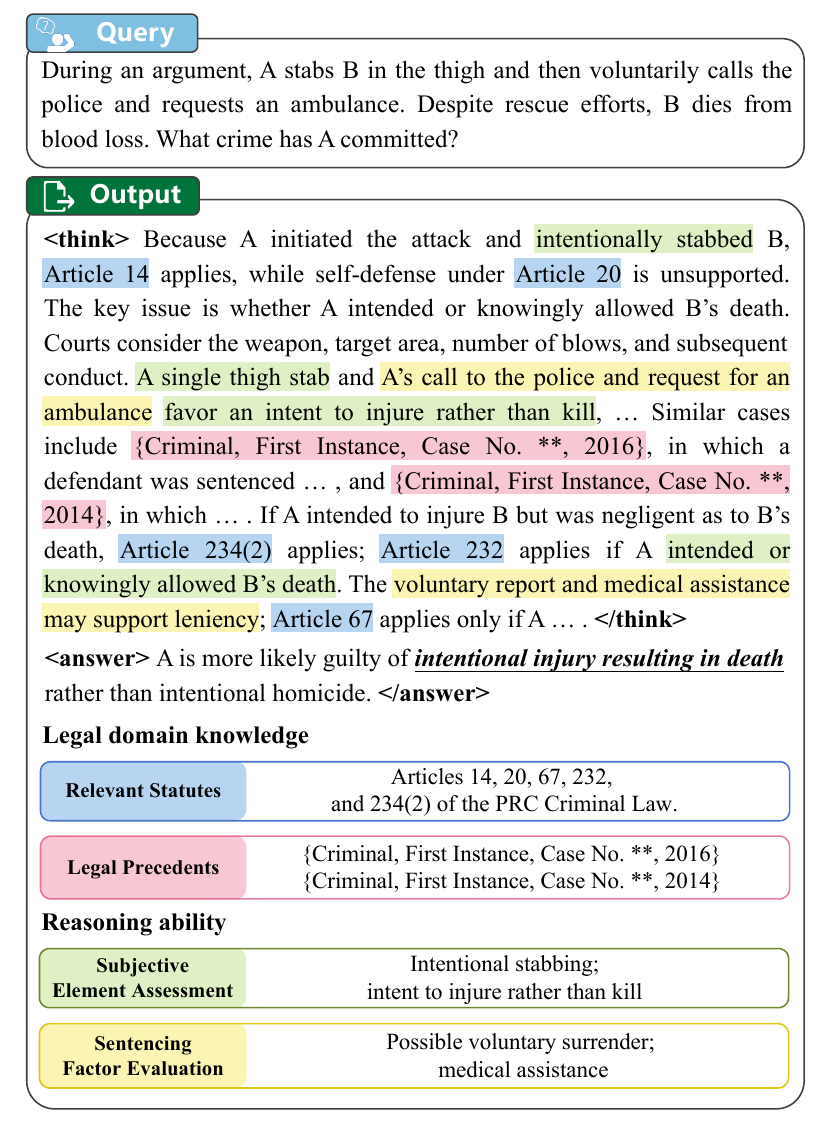}
    \caption{An anonymized and abridged example of a legal reasoning trace.}
    \label{fig:introduction}
\end{figure}

Recent advances in large language models (LLMs) have highlighted the importance of explicit reasoning, in which models generate intermediate thought processes before producing final answers, for enhancing reliability and interpretability across complex tasks. This form of structured reasoning is particularly crucial in the legal domain, where decisions must be grounded in legal principles and statutory provisions and must be logically coherent~\cite{levi2013introduction}.

We observe that high-quality legal reasoning traces are generally quite long, as they must encompass both necessary legal knowledge and a logically structured analysis, as shown in Figure~\ref{fig:introduction}. This requires a deep understanding of legal principles, including identifying relevant statutes and legal precedents, as well as the ability to accurately interpret and apply them~\cite{10.1093/acprof:oso/9780195150506.001.0001, li-etal-2025-unilr}. At the same time, legal reasoning must adhere to several critical logical constraints, including consistency between statutes and conclusions and internal logical coherence throughout the analysis~\cite{desai2016consistency}. Given the extensive domain-specific knowledge and rigorous logical reasoning required, together with the low tolerance for hallucinations in legal contexts~\cite{dahl2024large}, generating reliable legal reasoning traces remains significantly more challenging than in general domains.

Directly post-training a foundation model on legal-domain data is insufficient to address these challenges. During Supervised Fine-Tuning (SFT), reasoning traces are often several times as long as the final answers, causing token-level loss to bias optimization toward the traces and weaken the learning signal for final answers. Moreover, when reasoning traces contain flaws, such as misapplied precedents, SFT cannot provide targeted feedback to correct such errors. In reinforcement learning, methods such as Group Relative Policy Optimization (GRPO) have become mainstream alignment approaches. However, GRPO relies on outcome-level rewards that evaluate only final answers and ignore the legal soundness of intermediate reasoning. A straightforward solution is to introduce multiple rewards for different aspects of reasoning~\cite{cai-etal-2025-unilaw,wu-2025-comprehensive,wang2025answerthinkmultidimensionalsupervision}. However, this requires multiple manually designed reward functions, which are costly to construct, may introduce conflicting optimization objectives, gradually lose effectiveness during training, or even lead to reward hacking~\cite{pmlr-v235-wang24ay,laidlaw2025correlated}.

These challenges motivate our two-stage training framework for legal reasoning models. In Stage I, we introduce \textbf{JurisCoT-SFT}, which not only injects domain knowledge into the model but also adopts a length-normalized dual objective to mitigate signal imbalance. In Stage II, we propose \textbf{LABPO} (\textbf{L}ifecycle-\textbf{A}ware \textbf{B}acktrackable \textbf{P}olicy \textbf{O}ptimization), a self-regulating reinforcement learning framework for automatically managing reward signals during policy optimization.

The central idea of LABPO is to explicitly separate reward signals into two roles: \textbf{\textit{primary rewards}}, which directly correspond to the final evaluation objectives (e.g., F1 score and accuracy), and \textbf{\textit{auxiliary rewards}}, which serve as stage-dependent training signals (e.g., output format and language style). Unlike conventional approaches that keep all reward components active throughout training, LABPO treats auxiliary rewards as adaptive optimization instruments whose usefulness evolves over the course of learning. Each auxiliary signal remains active only while it remains beneficial to training; otherwise, it is retired, and the complete training state is rolled back to the best checkpoint observed during its active period.

We evaluate LLMs on selected tasks from three legal benchmarks: the statutory-law benchmarks ChJGBench~\cite{hu_2026_22171576} and LawBench~\cite{fei-etal-2024-lawbench}, and the case-law benchmark LegalBench~\cite{guha2023legalbench}. Using Qwen3-8B as the base model, our two-stage framework substantially improves legal reasoning performance across these benchmarks. On ChJGBench, it increases the average score from 0.6406 to 0.7786, achieving the highest score among all evaluated models and outperforming strong general-purpose LLMs such as Gemini-3-Pro-Preview. On LawBench, the resulting model achieves 0.848 on \textit{AP}, ranking second overall, and 0.666 on \textit{CP}, ranking first. It also performs strongly on case-law reasoning tasks in LegalBench, attaining the highest \textit{LRC} accuracy of 0.964 and competitive performance on the numerical reasoning task \textit{Sara-N}. Despite having only 8 billion parameters, the resulting Qwen3-8B-based model remains competitive with much larger models and even outperforms them on some tasks. Applying the same framework to Llama-3.1-8B likewise yields substantial gains, increasing its ChJGBench average score from 0.3374 to 0.6452 and improving its performance across LawBench and LegalBench. Ablation studies further demonstrate that both stages of the framework contribute to the observed performance improvements.

Our contributions can be summarized as follows:
\begin{enumerate}
\item We propose \textbf{JurisCoT-SFT}, which balances learning signals across reasoning traces and final answers, improving the model's ability to internalize legal knowledge and produce higher-quality reasoning outputs.
    
\item We introduce \textbf{LABPO}, which automatically manages the lifecycle of reward signals during policy optimization by dynamically evaluating and retiring auxiliary rewards based on their impact on final evaluation objectives.
    
\item We train a lightweight LLM that performs competitively across multiple benchmarks, establishing a strong lightweight baseline for future research on legal LLMs.
\end{enumerate}

\begin{figure*}[t]
    \centering
    \includegraphics[width=\textwidth, trim=5 10 5 10, clip]{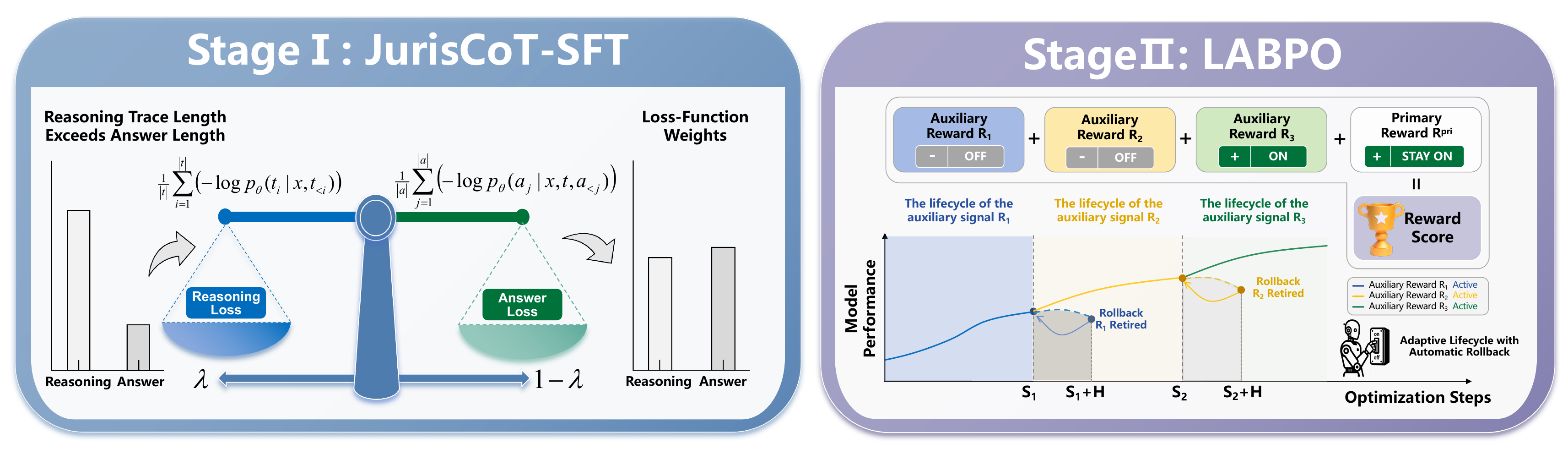}
    \caption{The framework has two stages. In Stage I, $\lambda$ balances learning between reasoning traces and final answers. In Stage II, auxiliary reward signals are activated, evaluated, and retired, with checkpoint rollback when needed.}
    \label{fig:2stage}
\end{figure*}

\section{Related Work}

Traditional legal-domain LLMs typically follow a two-stage paradigm consisting of continued pretraining and domain-specific fine-tuning to enhance legal capabilities. Representative models include LawGPT~\cite{zhou2024lawgptchineselegalknowledgeenhanced}, HanFei~\cite{HanFei}, LawLLM (US)~\cite{shu2024lawllm}, InternLM-Law~\cite{fei-etal-2025-internlm}, and WisdomInterrogatory~\cite{wu2026wisdominterrogatoryluwenopensourcelegal}. Previous studies have also shown that explicitly generating intermediate reasoning traces before producing final answers can improve performance on legal reasoning tasks~\cite{hu-etal-2025-evaluating}. DISC-LawLLM~\cite{yue2024lawllm} incorporates a legal syllogism framework into training data design. Recent work has further introduced reinforcement learning to improve reasoning quality in legal LLMs. LexPam~\cite{zhang-etal-2025-legal} focuses on legal numerical reasoning using a two-stage GRPO approach, optimizing numerical precision while incorporating legal constraints into the reward function. SyLeR~\cite{zhang2025SyLeR} uses supervised fine-tuning (SFT) as a warm-up stage, followed by Proximal Policy Optimization (PPO) to explore multiple reasoning paths, though its reward function mainly relies on general similarity metrics such as ROUGE, which may not fully capture the nuances of legal reasoning. LexPro~\cite{chen2025lexpro10technicalreport} integrates SFT and GRPO with rule-based format and process rewards, but these rewards primarily target output format and reasoning structure rather than domain-specific legal correctness. Legal$\Delta$~\cite{11464256} improves reasoning through chain-of-thought-guided information gain with a multidimensional reward mechanism. Despite recent advances, challenges remain in learning from long reasoning traces and in designing reward functions that effectively optimize reasoning performance. To address these challenges, we propose a two-stage training framework for legal reasoning models.

\section{Method}

We introduce a two-stage training framework for signal rebalancing and adaptive reward scheduling, as illustrated in Figure~\ref{fig:2stage}.

\subsection{Stage I: JurisCoT-SFT}

In the legal domain, reliable LLMs must ensure both procedural legitimacy and conclusion correctness. This requires a think-then-answer paradigm. However, using standard SFT presents a critical challenge: \textbf{the severe length imbalance between the typically lengthy reasoning traces and the concise final answers}. During training, this imbalance leads to undue emphasis on the "reasoning" part, weakening the learning signal for the shorter but crucial "answering" part. Given that even minor inaccuracies in the final answer can have high-stakes consequences, addressing this training imbalance is particularly important.

To address this issue, we propose \textbf{JurisCoT-SFT}, a variant of standard SFT that introduces a length-normalized dual objective. Formally, let a training instance be $(x, t, a)$, where $x$ denotes the input query, $t = (t_1, \dots, t_{|t|})$ represents the provided reasoning sequence, and $a = (a_1, \dots, a_{|a|})$ denotes the final answer. Given model parameters $\theta$, we optimize the next-token cross-entropy loss, with the reasoning and answer segments weighted separately and normalized by their respective lengths:

{\small
\begin{equation}
\begin{aligned}
&\mathcal{L}_{\text{JurisCoT-SFT}}(\theta) =
\lambda \cdot \tfrac{1}{|t|} \sum_{i=1}^{|t|}
\big(-\log p_\theta(t_i \mid x, t_{<i})\big) \\
&\quad + (1-\lambda) \cdot \tfrac{1}{|a|} \sum_{j=1}^{|a|}
\big(-\log p_\theta(a_j \mid x, t, a_{<j})\big),
\end{aligned}
\end{equation}
}

where $\lambda$ is a balancing coefficient that controls the relative contributions of the reasoning trace ($t$) and the final answer ($a$) to the training objective. A smaller \(\lambda\) places greater emphasis on the final-answer loss while retaining supervision from the reasoning trace.

\subsection{Stage II: Lifecycle-Aware Backtrackable Policy Optimization}
\label{sec:stage2_Adaptive_reward_grpo}

\begin{table}[t]
\centering
\small
\renewcommand{\arraystretch}{1.06}
\begin{tabular}{@{}c@{\hspace{0.75em}}l@{}}
\toprule
\textbf{Signal} & \textbf{Description} \\
\midrule
$S_1$ & Tag Format Validation \\
$S_2$ & JSON Schema Validation \\
$S_3$ & Statute Citation Format Verification \\
$S_4$ & Language Style Validation \\
$S_5$ & Statute Authenticity Verification \\
$S_6$ & Statute-Conclusion Consistency Verification \\
\bottomrule
\end{tabular}
\caption{Auxiliary reward signals used during training. Detailed definitions are provided in Appendix~\ref{sec:RewardFunctionforAuxiliarySignals}.}
\label{tab:aux_signals}
\end{table}

Although models fine-tuned with JurisCoT-SFT demonstrate strong legal knowledge and produce structured outputs, supervised fine-tuning cannot provide fine-grained supervision over intermediate reasoning steps or deliver differentiated feedback when reasoning traces contain errors.

To address these challenges, we propose \textbf{LABPO}, a reinforcement learning framework that \textbf{automatically manages reward signals} during policy optimization. The central idea of LABPO is to decompose reward signals into two functional roles. \textbf{Primary rewards} are aligned with the final evaluation objectives and remain active throughout training. \textbf{Auxiliary rewards} serve as stage-dependent training signals that may facilitate optimization during certain phases of learning but are not necessarily beneficial throughout the entire training trajectory. Instead of treating auxiliary rewards as fixed components of the reward function, LABPO assigns each auxiliary reward an independent lifecycle: after an initial monitoring period, it remains active as long as the running-best development score improves by at least $\epsilon$ over the most recent $H$ evaluation intervals and is retired otherwise.

Let $\tau$ denote the development-set evaluation index, with reward gates fixed between evaluations. For input $x$ and output $y$, the training reward is defined as:

{\small
\begin{equation}
\label{eq:primary_aux_total}
R(y;\tau) = R^{\text{pri}}(y)
+ \sum_{i=1}^{m} \alpha_i(\tau)\,R^{\text{aux}}_i(y),
\end{equation}
}

where \(R^{\mathrm{pri}}\) denotes the primary reward, \(\{R_i^{\mathrm{aux}}\}_{i=1}^{m}\) denotes the set of auxiliary rewards, and \(\alpha_i(\tau)\in\{0,1\}\) is a gating variable indicating whether the \(i\)-th auxiliary reward is active.

\subsubsection{Adaptive Auxiliary Signal Scheduling}
\label{sec:AASS}

To operationalize the lifecycle management of auxiliary rewards, we propose an Adaptive Auxiliary Signal Scheduling mechanism that automatically regulates the activation and retirement of auxiliary signals during training, thereby alleviating optimization plateaus and suppressing reward hacking. Pseudocode for this mechanism is provided in Appendix~\ref{sec:adaptive_signal_scheduling}.

\paragraph{Design Principles.} Instead of keeping all auxiliary signals active throughout the entire training process, we allow for the dynamic activation of auxiliary signals. To avoid interference between multiple auxiliary signals, only one auxiliary signal is active at any given time.

\paragraph{Periodic Evaluation.} Whenever the retained-step counter reaches a multiple of \(n\), we evaluate the current policy $\pi_{\theta_\tau}$ on a held-out development set $\mathcal{D}_{\text{dev}}$, which is independent of the test sets. The resulting performance score $S(\theta_\tau)$ serves as the sole criterion for determining the lifecycle of the auxiliary signal.

\paragraph{Dynamic Update of Auxiliary Signal Weights.} Let \(\tau_a\) denote the evaluation index at which the current auxiliary signal \(R^{\text{aux}}_a\) is activated. At evaluation step \(\tau\), we first define the running-best development score over the active interval:

{\small
\begin{equation}
B_\tau
=
\max_{j \in \{\tau_a,\ldots,\tau\}} S(\theta_j).
\label{eq:running_best_score}
\end{equation}
}

If multiple checkpoints attain the same best score \(B_\tau\), we select the latest one:

{\small
\begin{equation}
\tau^\star_\tau
=
\max\!\left\{
j \in \{\tau_a,\ldots,\tau\}
\,\middle|\,
S(\theta_j)=B_\tau
\right\}.
\label{eq:latest_best_checkpoint}
\end{equation}
}

During the first \(H\) evaluations after activation, the auxiliary signal remains active, defining a minimum active period before it becomes eligible for deactivation. Once at least \(H\) evaluations have been collected, we measure the improvement in the running-best score over the most recent \(H\) evaluation intervals:

{\small
\begin{equation}
G_\tau = B_\tau - B_{\tau-H},
\qquad
\tau - \tau_a \ge H.
\label{eq:window_gain}
\end{equation}
}

The weight of the active auxiliary signal is then updated as follows:

{\small
\begin{equation}
\alpha_a(\tau+1) =
\begin{cases}
0, & \tau - \tau_a \ge H \ \land\ G_\tau < \epsilon, \\
1, & \text{otherwise}.
\end{cases}
\label{eq:auxiliary_signal_update}
\end{equation}
}

The auxiliary signal is therefore retired only when the cumulative improvement in the running-best development score over the preceding \(H\) evaluation intervals is less than \(\epsilon\). Using the running-best score rather than the latest score makes the retirement decision less sensitive to short-term fluctuations during reinforcement learning.

\paragraph{Checkpoint Rollback and Mechanism Iteration.} Let \(\omega\), \(\zeta\), and \(s\) denote the optimizer state, learning-rate scheduler state, and retained-step counter, respectively. Each checkpoint stores \((\theta,\omega,\zeta,s)\) together with its associated development score. When the current auxiliary signal is retired, the complete training state is restored to the highest-scoring checkpoint observed during that signal's lifecycle:

{\small
\begin{equation}
\begin{aligned}
(\theta,\omega,\zeta,s) \leftarrow{}&
(\theta_{\tau^\star_\tau},\omega_{\tau^\star_\tau},
\zeta_{\tau^\star_\tau},s_{\tau^\star_\tau}).
\end{aligned}
\label{eq:best_checkpoint_rollback}
\end{equation}
}

The stored development score is reused to initialize the next phase, while the auxiliary-reward gates and phase variable are updated separately and are not restored from the checkpoint. The mechanism then activates the next auxiliary signal in the predetermined order and starts a new evaluation-and-scheduling cycle. After all auxiliary signals have been retired, all auxiliary rewards are disabled, and training continues using only the primary reward. Training ends when the retained-step counter reaches the target budget \(N\). If the running-best development score improves by less than \(\epsilon\) over \(H\) consecutive evaluation intervals before the budget is reached, the best checkpoint observed during this phase is restored and training terminates immediately.

\subsubsection{Auxiliary Reward Signal Integration}
\label{sec:AuxiliaryRewardSignalIntegration}

The goal of LABPO is not to eliminate reward design, but to make reward signals more reusable and easier to deploy. Unlike conventional reinforcement learning approaches that require carefully crafted reward functions to remain consistently effective throughout training, LABPO allows researchers to explore multiple auxiliary reward signals more freely, including potentially noisy or unstable heuristic signals, without pre-designing, tuning, or manually combining them in advance. Through dynamic evaluation during training, LABPO automatically identifies useful signals, retires those that cease to provide sufficient improvement, and rolls back the training state when necessary.

As shown in Table~\ref{tab:aux_signals}, we introduce six auxiliary reward signals based on prior work and common evaluation criteria for legal reasoning. The rationale for selecting these signals and their detailed definitions are provided in Appendix~\ref{sec:RewardFunctionforAuxiliarySignals}. In all main experiments, auxiliary signals are activated sequentially from $S_1$ to $S_6$. To assess whether this ordering affects final performance, we conduct an auxiliary-signal ordering experiment with the six proposed signals, as detailed in Appendix~\ref{sec:aux_signal_order}. The results show that, for the auxiliary signals considered in our framework, the optimization order has only a limited impact on final performance.

\begin{table*}[t]
\centering
\small
\setlength{\tabcolsep}{3.5pt}
\renewcommand{\arraystretch}{0.95}
\resizebox{\textwidth}{!}{
\begin{tabular}{@{}lcccccccccccc@{}}
\toprule
\multicolumn{1}{c}{\multirow{2.5}{*}{\textbf{Models}}} 
& \multicolumn{1}{c}{\multirow{2.5}{*}{\textbf{Avg.}}} 
& \multicolumn{1}{c}{\multirow{2.5}{*}{\textbf{Law}}} 
& \multicolumn{1}{c}{\multirow{2.5}{*}{\textbf{ISS}}} 
& \multicolumn{1}{c}{\multirow{2.5}{*}{\textbf{JUS}}} 
& \multicolumn{5}{c}{\textbf{Criminal Case}} 
& \multicolumn{2}{c}{\textbf{Civil Case}} 
& \multicolumn{1}{c}{\textbf{Admin.}} \\
\cmidrule(lr){6-10} \cmidrule(lr){11-12} \cmidrule(lr){13-13}
& & & & 
& \textbf{Fine}
& \textbf{MIT} 
& \textbf{Charge} 
& \textbf{Pen\_M} 
& \textbf{REC} 
& \textbf{SUP} 
& \textbf{PAY} 
& \textbf{DIS} \\
\midrule

\multicolumn{13}{c}{\textit{\underline{Specialized Legal LLMs}}} \\
\addlinespace[2.5pt]
WisdomInterrogatory  & 0.3728 & 0.158 & 0.582 & 0.384 & 0.135 & 0.408 & 0.436 & 0.608 & 0.108 & 0.649 & 0.244 & 0.326 \\
DISC-LawLLM-7B  & 0.4482 & 0.181 & 0.729 & 0.546 & 0.080 & 0.727 & 0.414 & 0.466 & 0.370 & 0.670 & 0.056 & 0.420 \\
\midrule

\multicolumn{13}{c}{\textit{\underline{Open-source LLMs}}} \\
\addlinespace[2.5pt]
Llama-3.1-405B  & 0.6006 & 0.342 & 0.712 & 0.575 & 0.476 & 0.885 & 0.957 & 0.602 & 0.500 & 0.682 & 0.616 & 0.712 \\
DeepSeek-R1  & 0.7080 & 0.422 & 0.819 & 0.718 & 0.563 & 0.961 & 0.969 & 0.740 & 0.715 & 0.783 & \cellcolor{toptwo}\underline{0.753} & 0.787 \\
Qwen3-235B-A22B-Instruct  & 0.7136 & 0.461 & 0.812 & \cellcolor{toptwo}\underline{0.748} & \cellcolor{toptwo}\underline{0.603} & \cellcolor{toptwo}\underline{0.974} & 0.967 & \cellcolor{toptwo}\underline{0.780} & \cellcolor{toptwo}\underline{0.765} & 0.778 & 0.744 & 0.653 \\
\midrule

\multicolumn{13}{c}{\textit{\underline{Proprietary LLMs}}} \\
\addlinespace[2.5pt]
Grok-4  & 0.7166 & 0.499 & 0.818 & 0.728 & 0.501 & 0.973 & \cellcolor{topone}\textbf{0.976} & 0.724 & 0.668 & 0.761 & 0.709 & 0.799 \\
GPT-5.1  & 0.6795 & 0.422 & 0.763 & 0.707 & 0.556 & 0.965 & \cellcolor{toptwo}\underline{0.975} & 0.691 & 0.665 & 0.764 & 0.722 & 0.687 \\
Claude-Sonnet-4.5  & 0.7293 & 0.540 & 0.811 & 0.729 & 0.577 & 0.968 & 0.970 & 0.751 & 0.705 & 0.779 & 0.743 & 0.764 \\
Gemini-3-Pro-Preview  & 0.7470 & 0.644 & 0.810 & 0.683 & \cellcolor{topone}\textbf{0.611} & 0.973 & 0.971 & \cellcolor{topone}\textbf{0.786} & 0.753 & \cellcolor{topone}\textbf{0.831} & \cellcolor{topone}\textbf{0.757} & 0.731 \\
\midrule

\multicolumn{13}{c}{\textit{\underline{Our Methods}}} \\
\addlinespace[2.5pt]

Llama-3.1-8B  & 0.3374 & 0.099 & 0.592 & 0.357 & 0.102 & 0.507 & 0.221 & 0.513 & 0.318 & 0.575 & 0.117 & 0.254 \\
with JurisCoT-SFT         & 0.6212 & 0.414 & 0.722 & 0.594 & 0.493 & 0.877 & 0.862 & 0.718 & 0.656 & 0.708 & 0.648 & 0.587 \\
with JurisCoT-SFT + LABPO & 0.6452 & 0.437 & 0.743 & 0.621 & 0.518 & 0.891 & 0.893 & 0.734 & 0.672 & 0.726 & 0.671 & 0.628 \\
\midrule
Qwen3-8B  & 0.6406 & 0.315 & 0.791 & 0.652 & 0.459 & 0.925 & 0.960 & 0.727 & 0.673 & 0.726 & 0.708 & 0.624 \\
with JurisCoT-SFT  & \cellcolor{toptwo}\underline{0.7667} & \cellcolor{toptwo}\underline{0.667} & \cellcolor{toptwo}\underline{0.830} & \cellcolor{toptwo}\underline{0.748} & 0.574 & \cellcolor{topone}\textbf{0.976} & \cellcolor{toptwo}\underline{0.975} & 0.770 & 0.747 & 0.801 & 0.717 & \cellcolor{toptwo}\underline{0.802} \\
with JurisCoT-SFT + LABPO  & \cellcolor{topone}\textbf{0.7786} & \cellcolor{topone}\textbf{0.689} & \cellcolor{topone}\textbf{0.846} & \cellcolor{topone}\textbf{0.751} & 0.579 & 0.971 & 0.974 & 0.766 & \cellcolor{topone}\textbf{0.784} & \cellcolor{toptwo}\underline{0.822} & 0.724 & \cellcolor{topone}\textbf{0.812} \\

\bottomrule
\end{tabular}
}
\caption{Evaluation results of different models on ChJGBench. The best results among all models are shown in \textbf{bold}, and the second-best results are \underline{underlined}.}
\label{tab:main_results_cn}
\end{table*}

\section{Experiments}

In this section, we describe the experimental setup, report the main results, analyze the training dynamics of auxiliary signal scheduling, compare reasoning performance, and conduct ablation studies. We additionally present a controlled diagnostic experiment on reward-hacking behavior in Appendix~\ref{sec:reward_hacking_diagnostic}.

\subsection{Experimental Setup}
\label{sec:Experimental_Setup}

\paragraph{Training Data.} We prepare training data for the LLMs in both statutory law and case law domains. For \textbf{statutory law}, we construct a large-scale legal training corpus consisting of \{Query, Reasoning Trace, Answer\} triplets. The query and answer are extracted from judicial documents, while the reasoning traces are generated by DeepSeek-R1. To ensure generation quality, we apply a variant of rejection sampling to filter generated traces (Appendix~\ref{sec:AttemptBoundedSelfVerification}). The dataset contains 2.73M instances (Table~\ref{tab:statute_corpus_stats}). For \textbf{case law}, we construct a corresponding dataset of triplets. The same rejection sampling procedure is applied to ensure the quality of the generated reasoning traces. The dataset contains 35,965 instances (Table~\ref{tab:case_corpus_stats}).

\paragraph{Evaluation Data.}

We evaluate LLMs on selected tasks from three legal benchmarks: ChJGBench~\cite{hu_2026_22171576} and LawBench~\cite{fei-etal-2024-lawbench} for statutory law, and LegalBench~\cite{guha2023legalbench} for case law. Task details are summarized in Table~\ref{tab:tasks_metrics}, and metric definitions are provided in Appendix~\ref{sec:EvaluationMetrics}.

\textbf{ChJGBench} is a benchmark comprising 3,462 real-world cases spanning criminal, civil, and administrative law. It evaluates three general dimensions shared by all cases: statutory provision application (\textit{Law}), disputed issues identification (\textit{ISS}), and judicial justification (\textit{JUS}). Domain-specific tasks include criminal charge prediction (\textit{Charge}), monetary fine (\textit{Fine}), recidivism assessment (\textit{REC}), mitigating circumstances identification (\textit{MIT}), penalty duration prediction (\textit{Pen\_M}), civil claim support and payment obligation (\textit{SUP}, \textit{PAY}), and administrative disposition prediction (\textit{DIS}).

\textbf{LawBench} is a public benchmark dataset for evaluating Chinese legal capabilities. We select three tasks: Article Prediction (\textit{AP}), Charge Prediction (\textit{CP}), and Sentence Prediction (\textit{SP}).

\textbf{LegalBench} is a benchmark for evaluating the legal reasoning abilities of LLMs using tasks derived from real legal datasets and expert-designed evaluations. We select three representative reasoning tasks for evaluation: Legal Reasoning Causality (\textit{LRC}), New York State Judicial Ethics (\textit{NYSJE}), and SARA Numeric Reasoning (\textit{Sara-N}).

\paragraph{Baselines.} We evaluate our model against a series of strong baselines across the benchmark test sets. These baselines include open-source general-purpose LLMs, such as Llama-3.1-405B~\cite{grattafiori2024llama3herdmodels}, DeepSeek-R1~\cite{Guo_2025} and Qwen3-235B-A22B-Instruct~\cite{yang2025qwen3technicalreport}; closed-source general-purpose LLMs, including Grok-4~\cite{xai2025grok4}, GPT-5.1~\cite{openai_gpt5.1_systemcard_2025}, Claude-Sonnet-4.5~\cite{anthropic2025claudesonnet45} and Gemini-3-Pro-Preview~\cite{googledeepmind2026gemini3pro}; and legal-domain-specific LLMs, including WisdomInterrogatory~\cite{wu2026wisdominterrogatoryluwenopensourcelegal} and DISC-LawLLM-7B~\cite{yue2024lawllm}.

\paragraph{Implementation Details.}
We use Llama-3.1-8B and Qwen3-8B as the base models for subsequent fine-tuning. Since Qwen3-8B achieves better overall performance, we conduct all ablation studies using Qwen3-8B as the backbone. Detailed training hyperparameters are provided in Appendix~\ref{sec:implementation_details}, and the computational overhead of LABPO is reported in Appendix~\ref{sec:ComputationalOverheadAnalysis}.

\subsection{Main Results}

\subsubsection{Results on Statutory-Law Benchmarks}

\begin{table}[t]
\centering
\small
\setlength{\tabcolsep}{5pt}
\renewcommand{\arraystretch}{0.95}
\begin{tabular}{lccc}
\toprule
\textbf{Models} & \textbf{AP} $\uparrow$ & \textbf{CP} $\uparrow$ & \textbf{SP} $\uparrow$\\
\midrule

\multicolumn{4}{c}{\textit{\underline{Specialized Legal LLMs}}} \\
\addlinespace[2.5pt]
WisdomInterrogatory  & 0.328 & 0.351 & 0.804 \\
DISC-LawLLM-7B               & 0.054 & 0.127 & 0.670 \\
\midrule

\multicolumn{4}{c}{\textit{\underline{Open-source LLMs}}} \\
\addlinespace[2.5pt]
Llama-3.1-405B       & 0.306 & 0.352 & 0.807 \\
DeepSeek-R1          & 0.831 & 0.606 & \cellcolor{topone}\textbf{0.891} \\
Qwen3-235B-A22B-Instruct & \cellcolor{topone}\textbf{0.865} & 0.607 & 0.857 \\

\midrule

\multicolumn{4}{c}{\textit{\underline{Proprietary LLMs}}} \\
\addlinespace[2.5pt]
Grok-4               & 0.844 & 0.553 & \cellcolor{toptwo}\underline{0.885} \\
GPT-5.1              & 0.721 & 0.561 & 0.883 \\
Claude-Sonnet-4.5    & 0.818 & 0.550 & 0.775 \\
Gemini-3-Pro-Preview & 0.779 & \cellcolor{toptwo}\underline{0.637} & 0.822 \\
\midrule

\multicolumn{4}{c}{\textit{\underline{Our Methods}}} \\
\addlinespace[2.5pt]
Llama-3.1-8B       & 0.057 & 0.223 & 0.707 \\
with JurisCoT-SFT    & 0.756 & 0.254 & 0.849 \\
with JurisCoT-SFT + LABPO & 0.771 & 0.288 & 0.862 \\
\midrule
Qwen3-8B             & 0.792 & 0.537 & 0.837 \\
with JurisCoT-SFT    & 0.822 & 0.613 & 0.843 \\
with JurisCoT-SFT + LABPO & \cellcolor{toptwo}\underline{0.848} & \cellcolor{topone}\textbf{0.666} & 0.854 \\
\bottomrule
\end{tabular}

\caption{Evaluation results of different models on LawBench. \textbf{AP}, \textbf{CP}, and \textbf{SP} denote Article Prediction, Charge Prediction, and Sentence Prediction, respectively. The best results among all models are shown in \textbf{bold}, and the second-best results are \underline{underlined}.}
\label{tab:main_results_lawbench}
\end{table}

We report the main experimental results on the statutory-law benchmarks ChJGBench and LawBench in Tables~\ref{tab:main_results_cn} and~\ref{tab:main_results_lawbench}.

\paragraph{Specialized legal LLMs built on outdated base models show limited capability.}
Models designed specifically for the legal domain perform relatively poorly overall. On ChJGBench, WisdomInterrogatory and DISC-LawLLM-7B achieve average scores of only 0.3728 and 0.4482, respectively, far below most modern general-purpose LLMs and our trained models. Their weaknesses are particularly evident in decision-oriented tasks such as \textit{Fine} and \textit{PAY}, where both models obtain substantially lower scores than recent open-source and proprietary LLMs. A similar pattern is observed on LawBench, where both models substantially underperform most recent general-purpose LLMs on Article Prediction and Charge Prediction. These results suggest that legal-domain training alone may be insufficient when the underlying base models lack strong general reasoning and instruction-following capabilities.

\paragraph{General-purpose LLMs perform reasonably well overall but still exhibit limitations in legal knowledge.}
Modern general-purpose LLMs achieve relatively strong overall performance across both benchmarks, indicating that large-scale pretraining provides a useful foundation for legal reasoning. On ChJGBench, Gemini-3-Pro-Preview obtains the highest average score among general-purpose models at 0.7470. However, these models may still show weaker performance on tasks requiring precise grounding in legal knowledge. For example, Gemini-3-Pro-Preview achieves the highest \textit{Law} score among general-purpose models (0.644), but remains below our best-performing model (0.689).

\paragraph{Our two-stage training framework consistently improves lightweight base models.}
Applying our proposed two-stage framework consistently improves both lightweight backbones. On ChJGBench, Llama-3.1-8B improves from 0.3374 to 0.6452 after JurisCoT-SFT and LABPO, while Qwen3-8B improves from 0.6406 to 0.7786. The final Qwen3-8B-based model achieves the best overall score among all evaluated models and obtains top performance on key metrics such as \textit{Law} (0.689), \textit{ISS} (0.846), \textit{JUS} (0.751), \textit{REC} (0.784), and \textit{DIS} (0.812). A similar trend is observed on LawBench.

\subsubsection{Results on Case Law Benchmark}

\begin{table}[t]
\centering
\small
\setlength{\tabcolsep}{2.5pt}
\renewcommand{\arraystretch}{0.95}
\begin{tabular}{@{}lccc@{}}
\toprule
\textbf{Models} & \textbf{LRC} $\uparrow$ & \textbf{NYSJE} $\uparrow$ & \textbf{Sara-N} $\downarrow$\\
\midrule

\multicolumn{4}{c}{\textit{\underline{Specialized Legal LLMs}}} \\
\addlinespace[2.5pt]
WisdomInterrogatory  & 0.491 & 0.387 & -- \\
DISC-LawLLM-7B               & 0.746 & 0.630 & 19.330 \\
\midrule

\multicolumn{4}{c}{\textit{\underline{Open-source LLMs}}} \\
\addlinespace[2.5pt]
Llama-3.1-405B       & 0.909 & 0.709 & 7.720 \\
DeepSeek-R1          & 0.927 & 0.771 & \cellcolor{topone}\textbf{0.250} \\
Qwen3-235B-A22B-Instruct & 0.891 & 0.791 & 9.618 \\

\midrule

\multicolumn{4}{c}{\textit{\underline{Proprietary LLMs}}} \\
\addlinespace[2.5pt]
Grok-4               & 0.909 & \cellcolor{toptwo}\underline{0.805} & 5.802 \\
GPT-5.1              & \cellcolor{toptwo}\underline{0.946} & \cellcolor{topone}\textbf{0.836} & 7.051 \\
Claude-Sonnet-4.5    & 0.891 & 0.747 & \cellcolor{toptwo}\underline{0.994} \\
Gemini-3-Pro-Preview & 0.873 & 0.764 & 7.677 \\
\midrule

\multicolumn{4}{c}{\textit{\underline{Our Methods}}} \\
\addlinespace[2.5pt]
Llama-3.1-8B       & 0.764 & 0.493 & 21.088 \\
with JurisCoT-SFT    & 0.855 & 0.575 & 6.125 \\
with JurisCoT-SFT + LABPO & 0.909 & 0.623 & 4.846 \\
\midrule
Qwen3-8B             & 0.873 & 0.596 & 8.093 \\
with JurisCoT-SFT      & 0.927 & 0.627 & 3.949 \\
with JurisCoT-SFT + LABPO & \cellcolor{topone}\textbf{0.964} & 0.688 & 1.753 \\
\bottomrule
\end{tabular}

\caption{Evaluation results of different models on LegalBench. The best results among all models are shown in \textbf{bold}, and the second-best results are \underline{underlined}.}
\label{tab:main_results_en}
\end{table}

We report the main experimental results for three case-law tasks from LegalBench in Table~\ref{tab:main_results_en}. The results show that our two-stage framework generalizes beyond statutory-law tasks and remains effective across the evaluated case-law tasks. In particular, Qwen3-8B with JurisCoT-SFT + LABPO achieves the best performance on \textit{LRC}, improving from 0.873 to 0.964 and outperforming all evaluated models. It also substantially reduces the numerical error on \textit{Sara-N}, from 8.093 to 1.753. For the \textit{NYSJE} task, performance is limited by the model size, resulting in lower scores than those of larger proprietary models. However, it still demonstrates a significant improvement over the base model. Overall, these results demonstrate the effectiveness and cross-domain applicability of our training framework.

\subsection{Training Dynamics of Auxiliary Signal Scheduling}

\begin{figure}[h]
    \centering
    \includegraphics[
        width=1\linewidth, trim=20 15 10 15,clip
    ]{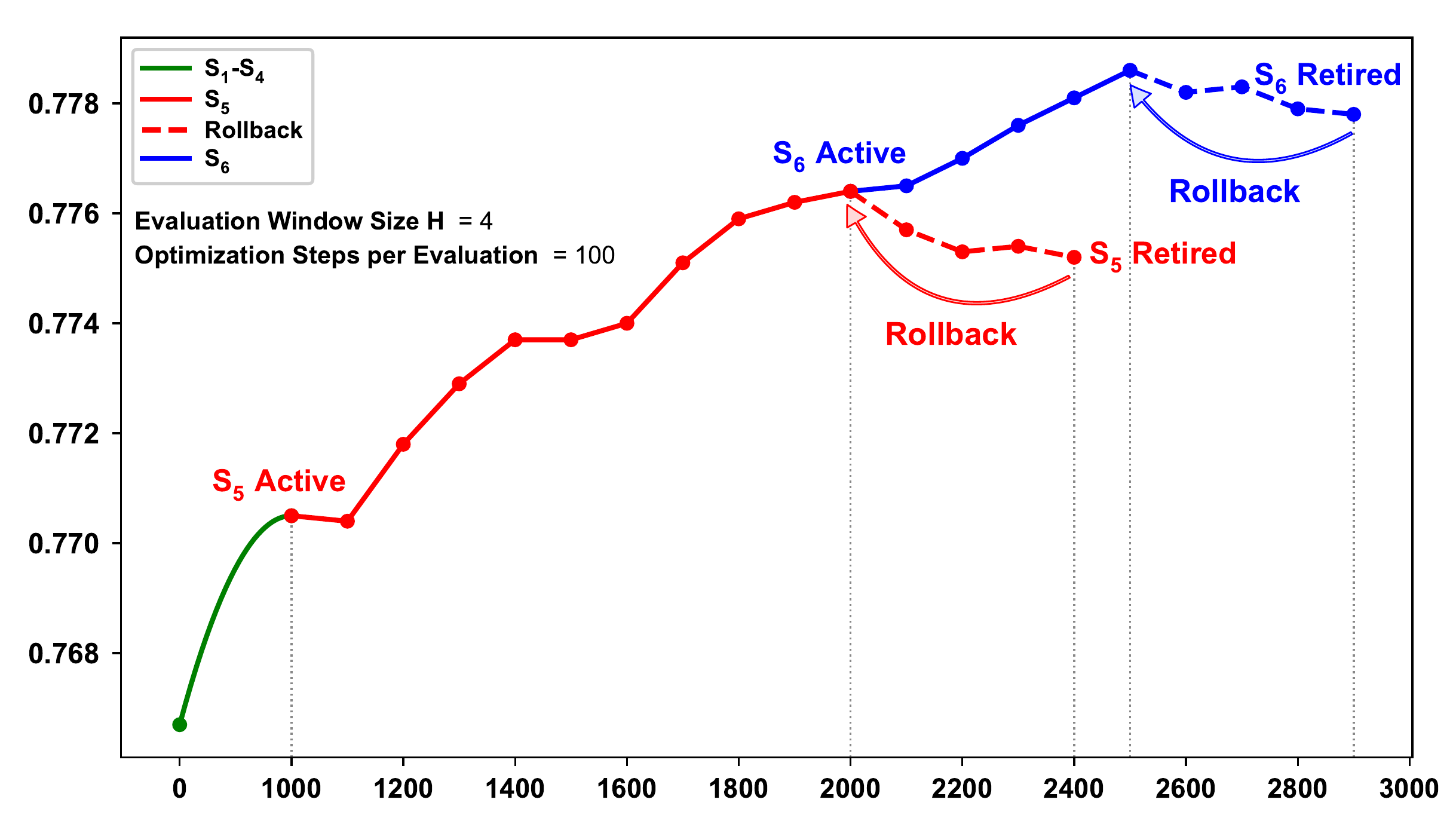}
    \caption{Training dynamics of auxiliary-signal scheduling in LABPO. Detailed trajectories for $S_1$--$S_4$ are omitted for clarity.}
    \label{fig:labpo_result}
\end{figure}

As shown in Figure~\ref{fig:labpo_result}, the auxiliary signal \(S_5\) initially improves the development score but later reaches a plateau. When the running-best development score improves by less than \(\epsilon\) over \(H=4\) consecutive evaluation intervals, \(S_5\) is retired and the policy is restored to the highest-scoring checkpoint observed since \(S_5\) was activated. The next auxiliary signal, \(S_6\), is then activated, after which development performance resumes an upward trend and eventually exceeds the previous peak obtained under \(S_5\). This trajectory indicates that an auxiliary reward may be useful during only part of the optimization process and may become ineffective as training progresses. Restoring the best checkpoint limits the performance degradation accumulated after the auxiliary signal ceases to provide sufficient improvement.

\subsection{Reasoning Performance Comparison}

\begin{figure}[t]
    \centering
    \includegraphics[width=\columnwidth, trim=5 5 5 5, clip]{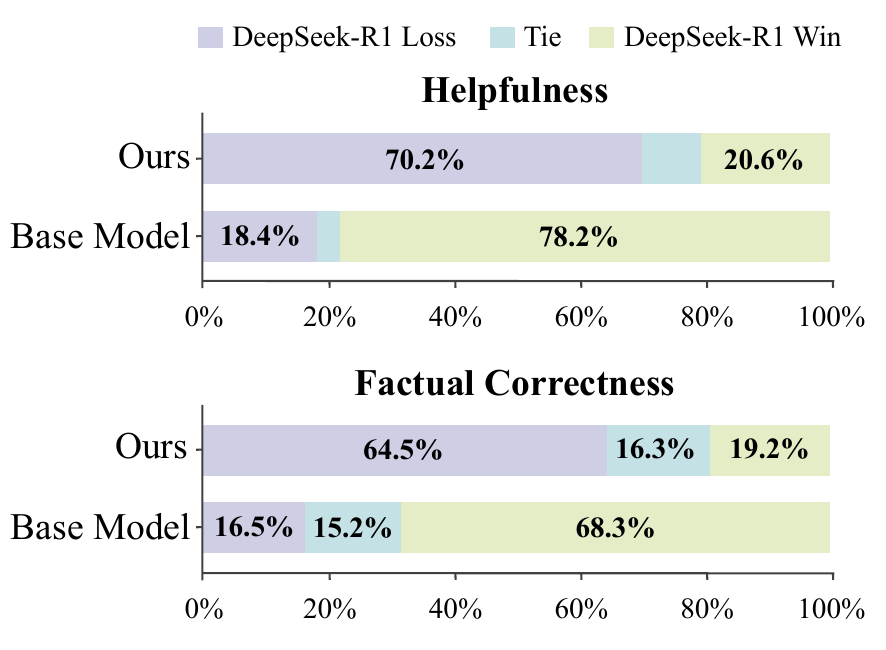}
    \caption{Pairwise reasoning-trace win rates against DeepSeek-R1 on ChJGBench. Ours: our two-stage model; Baseline: Qwen3-8B. Purple, blue, and green denote a DeepSeek-R1 loss, tie, and win, respectively.}
    \label{fig:WinRate}
\end{figure}

To further investigate improvements in reasoning ability, we conduct a direct evaluation of reasoning traces. Using a win-rate framework, we separately compare the model trained with our method and the Qwen3-8B baseline against DeepSeek-R1. We evaluate reasoning traces along two general dimensions: \textbf{Helpfulness} and \textbf{Factual Correctness}. Gemini-3-Pro-Preview is employed as the judge.

The details of this experiment can be found in Appendix~\ref{sec:ReasoningPerformanceComparisonDetails}. The specifics of the alignment evaluation between LLMs and human annotators are available in Appendix~\ref{sec:human_evaluation}. The LLM evaluator shows strong alignment with aggregated human judgments, achieving an accuracy of 0.84 and a Cohen's $\kappa$ of 0.756.

As shown in Figure~\ref{fig:WinRate}, our method demonstrates notable improvements in reasoning quality compared to the baseline Qwen3-8B. In direct comparisons with the strong open-source model DeepSeek-R1, the win rate of our approach increases from 18.37\% to 70.19\% on Helpfulness and from 16.46\% to 64.47\% on Factual Correctness. Overall, the proposed training strategy enables an 8B-scale model to achieve reasoning performance on legal tasks that is competitive with or even superior to that of the 671B-parameter MoE model DeepSeek-R1.

\subsection{Ablation Studies}

Our ablation study on ChJGBench (Table~\ref{tab:ablation_study}) evaluates each component's contribution; the trends on LegalBench and LawBench are consistent (see Appendix~\ref{sec:Ablation_Studies_on_LawBench_and_LegalBench}).

\paragraph{(1) LABPO Mechanisms Improve Final RL Performance.}
Rows 1-4 analyze the effect of the Retire and Rollback mechanisms during reinforcement learning (RL). The full configuration (Row 1) achieves the best performance of 0.7786. Removing Rollback (Row 2) decreases the score to 0.7718, and removing both mechanisms (Row 3) further decreases it to 0.7671, indicating that both core mechanisms contribute to LABPO's final performance. Moreover, Row 3 performs slightly worse than Row 4, which optimizes using only the primary reward. A possible reason is that different rewards may not share fully aligned optimization objectives, and aggregating multiple rewards can introduce noise or optimization conflicts.

\paragraph{(2) JurisCoT-SFT Improves the Effectiveness of Supervised Training.}
When JurisCoT-SFT is adopted (Row 5), the model achieves a score of 0.7667, outperforming naive SFT (Row 7, 0.7399). A further ablation study of the SFT objectives, together with a sensitivity analysis of $\lambda$, is provided in Appendix~\ref{app:sft_signal_rebalancing}.

\paragraph{(3) LABPO Improves Performance Across Initialization Settings.}
Comparisons of Rows 1/5, 6/7, and 8/9 show that LABPO consistently improves performance under different initializations, boosting JurisCoT-SFT, naive SFT, and base models from 0.7667 to 0.7786, 0.7399 to 0.7687, and 0.6406 to 0.6735, respectively. This indicates that the gains stem not only from JurisCoT-SFT initialization but also from the proposed RL strategy.

\begin{table}[t]
\centering
\small
\setlength{\tabcolsep}{3pt}
\renewcommand{\arraystretch}{0.95}
\begin{tabular}{c cc ccc r}
\toprule
\multirow{2}{*}{\textbf{No.}} 
& \multicolumn{2}{c}{\textbf{SFT}} 
& \multicolumn{3}{c}{\textbf{RL}} 
& \multirow{2}{*}{\textbf{Avg.} $\uparrow$} \\
\cmidrule(lr){2-3}
\cmidrule(lr){4-6}
& Used & Variant & Used & Retire & Rollback & \\
\midrule
1 & \checkmark & \checkmark & \checkmark & \checkmark & \checkmark & 0.7786 \\
2 & \checkmark & \checkmark & \checkmark & \checkmark & $\times$ & 0.7718 \\
3 & \checkmark & \checkmark & $\bullet$ & $\times$ & $\times$ & 0.7671 \\
4 & \checkmark & \checkmark & $\circ$ & $\times$ & $\times$ & 0.7684 \\
5 & \checkmark & \checkmark & $\times$ & $\times$ & $\times$ & 0.7667 \\
6 & \checkmark & $\times$ & \checkmark & \checkmark & \checkmark & 0.7687 \\
7 & \checkmark & $\times$ & $\times$ & $\times$ & $\times$ & 0.7399 \\
8 & $\times$ & $\times$ & \checkmark & \checkmark & \checkmark & 0.6735 \\
9 & $\times$ & $\times$ & $\times$ & $\times$ & $\times$ & 0.6406 \\
\bottomrule
\end{tabular}
\caption{Ablation study of different training configurations.
\checkmark indicates the component is enabled and $\times$ indicates it is disabled. Row 3 reduces to GRPO with aggregated primary and auxiliary rewards, while Row 4 reduces to GRPO using only the primary reward.}
\label{tab:ablation_study}
\end{table}

\section{Conclusion}
In this paper, we introduce a two-stage training framework to enhance the legal reasoning capabilities of LLMs. We propose \textbf{JurisCoT-SFT}, which balances learning signals between reasoning traces and final answers, and \textbf{LABPO}, which dynamically schedules auxiliary reward signals. Our approach achieves strong performance on statutory and case law benchmarks. The ablation results further confirm the contributions of both training stages.

\section*{Limitations}
Despite the strong empirical performance and methodological innovations of our approach, several limitations should be acknowledged. First, although the auxiliary signals in the LABPO framework no longer require meticulous selection thanks to the backtracking mechanism, the model still depends on a predefined reward function. Fully automating the design of reward functions (e.g., enabling LLMs to autonomously design appropriate reward signals) remains an open challenge, and we leave this direction for future work. Second, due to resource constraints, we were unable to validate the effectiveness of our proposed two-stage approach on larger-scale models. Third, compared with simpler reinforcement learning methods such as GRPO, LABPO introduces additional computational overhead. The lifecycle management of auxiliary reward signals, periodic evaluation on the development set, and potential rollback operations require extra training iterations and evaluation costs. Although these mechanisms contribute to final performance, they may increase the overall computational budget. Designing more efficient reward scheduling or lightweight rollback strategies to reduce this overhead is an important direction for future research.

\section*{Ethics Statement}
Given the sensitive nature of the legal domain, the application of artificial intelligence in this field necessitates rigorous ethical management. To address potential ethical concerns, we implement several safeguards. In particular, to prevent the leakage of private information (e.g., personal names), we replace sensitive information in the training datasets with neutral third-person references. This ensures that our research upholds the principles of privacy protection and responsible AI development.

\section*{Acknowledgements}

This work was supported by the "Pioneer" and "Leading Goose" R\&D Program of Zhejiang (2025C02037), Research Project of the Philosophy and Social Sciences Planning Research Platform (Laboratory) of Zhejiang Province, and Key R\&D Program of Hangzhou (2025SZDA0254). It was also supported by the National Natural Science Foundation of China (62106039) and Alibaba Research Intern Program. Any opinions expressed in this paper are those of the authors and do not necessarily reflect the views of the funding organizations.

We declare that AI-based writing tools were used only for limited language polishing and grammatical refinement in preparing this paper. The authors have carefully verified all AI-assisted text and take full responsibility for the content of the paper.

\bibliography{main}

\appendix

\section{Attempt-Bounded Self-Verification}
\label{sec:AttemptBoundedSelfVerification}

\begin{figure}[b]
    \centering
    \includegraphics[width=\columnwidth, trim=5 5 5 5, clip]{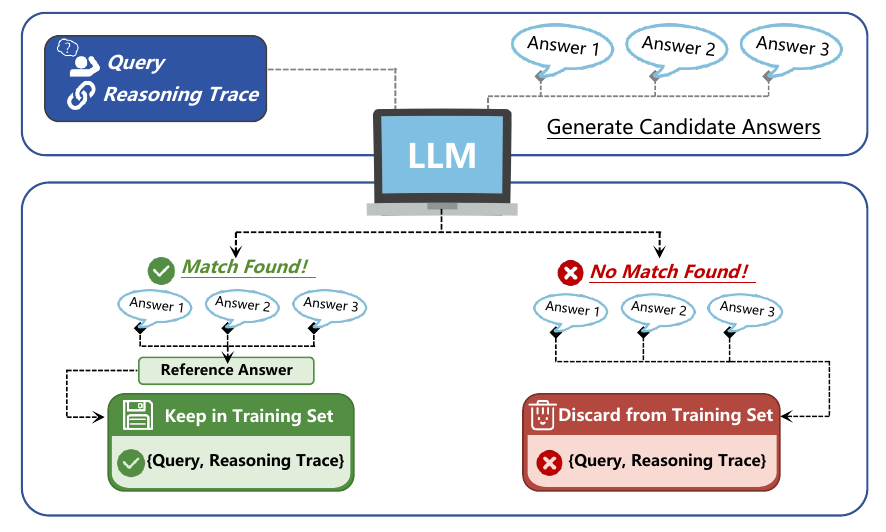}
    \caption{An overview of the Attempt-Bounded Self-Verification framework, in which a query and its reasoning trace are fed into the target model, and the pair is retained only if the model can infer the correct answer within a bounded number of attempts.}
    \label{fig:AttemptBoundedSelfVerification}
\end{figure}

This section presents a method for improving the quality of training data by filtering reasoning traces, as illustrated in Figure~\ref{fig:AttemptBoundedSelfVerification}. This method is called \textbf{Attempt-Bounded Self-Verification}. Specifically, given a question--answer pair \( (x, a) \), we prompt a strong LLM to generate a reasoning trace \( t \). We then feed the pair \( (x, t) \) into the model to be trained, which produces multiple candidate answers \( \hat{a}^{(k)} \). If any candidate answer matches the ground-truth answer \( a \), we regard the corresponding training triple \( (x, t, a) \) as valid and retain it in the training set. We allow up to \( K \) attempts; if none of the generated candidates matches the reference answer, the training triple is discarded.

{\small
\begin{equation}
\begin{aligned}
\mathcal{D}' = \Big\{ (x,t,a) \in \mathcal{D} \;\Big|\;
&\exists\, k \leq K,\ \hat{a}^{(k)} \sim p_\theta(\cdot \mid x,t), \\
&v(\hat{a}^{(k)},a) = 1
\Big\}.
\end{aligned}
\end{equation}
}

Here, \( \hat{a}^{(k)} \) denotes the \( k \)-th candidate answer sampled from the model conditioned on \( (x, t) \), 
\( p_\theta(\cdot \mid x, t) \) represents the conditional distribution defined by model parameters \( \theta \), 
and \( v(\hat{a}^{(k)}, a) \in \{0, 1\} \) is a verification function indicating whether the candidate answer agrees with the ground-truth answer \( a \).

\section{Pseudocode for Adaptive Auxiliary Signal Scheduling}
\label{sec:adaptive_signal_scheduling}

Algorithm~\ref{alg:LABPO} summarizes the scheduling procedure. Each checkpoint stores the policy parameters, optimizer state, learning-rate scheduler state, retained-step counter, and associated development score; \(\operatorname{score}(\mathcal{C})\) returns the score stored in checkpoint \(\mathcal{C}\).

\begin{algorithm*}[t]
\caption{Adaptive Auxiliary Signal Scheduling for Legal Reasoning}
\label{alg:LABPO}
\small
\begin{algorithmic}[1]

\Require Training set \(\mathcal{D}_t\); policy \(\pi_\theta\); optimizer state \(\omega\); learning-rate scheduler state \(\zeta\)

\Require Development set \(\mathcal{D}_{\mathrm{dev}}\); score \(S(\theta)\)

\Require Auxiliary rewards \(\{R_i^{\mathrm{aux}}\}_{i=1}^{m}\); activation order \(\boldsymbol{\sigma}=(\sigma_1,\ldots,\sigma_m)\)

\Require Target retained-step budget \(N\); evaluation interval \(n\); minimum gain \(\epsilon\); patience window \(H\)

\Ensure Final policy parameters \(\theta\)

\State \(\boldsymbol{\alpha}\gets\mathbf{0}\), \(k\gets1\), \(a\gets\sigma_k\), \(\alpha_a\gets1\), and \(\mathrm{phase}\gets\mathrm{auxiliary}\)

\State Set the retained-step counter \(s\gets0\) and the lifecycle-local evaluation index \(r\gets0\)

\State \(B_0\gets S(\theta)\); save
\(\mathcal{C}^{\mathrm{best}}\gets
\operatorname{Checkpoint}(\theta,\omega,\zeta,s,B_0)\)

\While{\(s<N\)}

    \State Sample \(x\sim\mathcal{D}_t\), roll out \(y\sim\pi_\theta(\cdot\mid x)\), and update \(\pi_\theta\) by GRPO using Eq.~\ref{eq:primary_aux_total}

    \State \(s\gets s+1\)

    \If{\(s\bmod n=0\)}

        \State \(r\gets r+1\); compute \(q_r\gets S(\theta)\)

        \If{\(q_r\geq B_{r-1}\)}
            \State \(\mathcal{C}^{\mathrm{best}}\gets
            \operatorname{Checkpoint}(\theta,\omega,\zeta,s,q_r)\)
        \EndIf

        \State \(B_r\gets\max(B_{r-1},q_r)\)

        \If{\(r\geq H\) \textbf{and} \(B_r-B_{r-H}<\epsilon\)}

            \If{\(\mathrm{phase}=\mathrm{auxiliary}\)}

                \State Retire the active reward: \(\alpha_a\gets0\)

                \State Restore \((\theta,\omega,\zeta,s)\) from \(\mathcal{C}^{\mathrm{best}}\); keep \(\boldsymbol{\alpha}\) and \(\mathrm{phase}\) unchanged

                \If{\(k<m\)}
                    \State \(k\gets k+1\), \(a\gets\sigma_k\), and \(\alpha_a\gets1\)
                \Else
                    \State \(\boldsymbol{\alpha}\gets\mathbf{0}\) and \(\mathrm{phase}\gets\mathrm{primary}\)
                \EndIf

                \State Reset the phase-local history:
                \(r\gets0\) and
                \(B_0\gets\operatorname{score}(\mathcal{C}^{\mathrm{best}})\)

            \Else

                \State Restore \((\theta,\omega,\zeta,s)\) from \(\mathcal{C}^{\mathrm{best}}\); keep \(\boldsymbol{\alpha}\) and \(\mathrm{phase}\) unchanged

                \State \Return \(\theta\)
                \Comment{Terminate after restoring the best checkpoint from the phase using only the primary reward}

            \EndIf

        \EndIf

    \EndIf

\EndWhile

\State \Return \(\theta\)
\Comment{The retained-step budget \(N\) has been reached}

\end{algorithmic}
\end{algorithm*}

\section{Effect of Auxiliary Signal Ordering}
\label{sec:aux_signal_order}

Because LABPO activates auxiliary signals sequentially, we examine its sensitivity to signal ordering. To isolate this factor, all runs start from the Qwen3-8B base checkpoint, bypass the JurisCoT-SFT stage, and are trained on the Chinese statutory-law dataset using LABPO. The six auxiliary signals defined in Table~\ref{tab:aux_signals} are kept fixed, with only their order varied across the following five settings:

\begin{enumerate}[topsep=2pt, partopsep=2pt, parsep=0pt, itemsep=2pt]
\item Forward order: {\small $S_1 \!\to\! S_2 \!\to\! S_3 \!\to\! S_4 \!\to\! S_5 \!\to\! S_6$}
\item Reverse order: {\small $S_6 \!\to\! S_5 \!\to\! S_4 \!\to\! S_3 \!\to\! S_2 \!\to\! S_1$}
\item Random order 1: {\small $S_4 \!\to\! S_3 \!\to\! S_6 \!\to\! S_1 \!\to\! S_5 \!\to\! S_2$}
\item Random order 2: {\small $S_5 \!\to\! S_1 \!\to\! S_2 \!\to\! S_4 \!\to\! S_3 \!\to\! S_6$}
\item Random order 3: {\small $S_3 \!\to\! S_4 \!\to\! S_1 \!\to\! S_6 \!\to\! S_5 \!\to\! S_2$}
\end{enumerate}

\begin{table}[t]
\centering
\small
\begin{tabular}{cc}
\toprule
\textbf{Signal Order} & \textbf{Avg.} $\uparrow$ \\
\midrule
$S_1 \rightarrow S_2 \rightarrow S_3 \rightarrow S_4 \rightarrow S_5 \rightarrow S_6$ & 0.6735 \\
$S_6 \rightarrow S_5 \rightarrow S_4 \rightarrow S_3 \rightarrow S_2 \rightarrow S_1$ & 0.6682 \\
$S_4 \rightarrow S_3 \rightarrow S_6 \rightarrow S_1 \rightarrow S_5 \rightarrow S_2$ & 0.6671 \\
$S_5 \rightarrow S_1 \rightarrow S_2 \rightarrow S_4 \rightarrow S_3 \rightarrow S_6$ & 0.6723 \\
$S_3 \rightarrow S_4 \rightarrow S_1 \rightarrow S_6 \rightarrow S_5 \rightarrow S_2$ & 0.6704 \\
\bottomrule
\end{tabular}
\caption{Ablation study on the effect of auxiliary signal ordering using Qwen3-8B. Reported values are average performance scores computed using Eq. (\ref{eq:weighted_avg_score}).}
\label{tab:signal_order_ablation}
\end{table}

Table~\ref{tab:signal_order_ablation} summarizes the results for the five tested signal orders. The average scores range from 0.6671 to 0.6735, with a mean of 0.6703 and a sample standard deviation of 0.0027. The difference between the highest and lowest scores is 0.0064. \textbf{These findings suggest limited performance variation across the tested signal orders in our experimental setting.} This conclusion is specific to the auxiliary signals evaluated here and should not be interpreted as evidence that auxiliary signals beyond this set are similarly insensitive to ordering.

\section{Reward Functions for Auxiliary Signals}
\label{sec:RewardFunctionforAuxiliarySignals}

Section~\ref{sec:AuxiliaryRewardSignalIntegration} and Table~\ref{tab:aux_signals} introduce six auxiliary reward signals for legal reasoning. This appendix provides their detailed definitions and calculation formulas.

These signals broadly cover three categories of errors. Tag Format Validation and JSON Schema Validation address output structure and parsability. Statute Citation Format Verification and Language Style Validation address legal expression and presentation. Statute Authenticity Verification and Statute-Conclusion Consistency Verification address substantive legal reliability.

\subsection{Tag Format Validation}
Model outputs are expected to follow a strict structure consisting of exactly one \texttt{<think>...</think>} block followed by exactly one \texttt{<answer>...</answer>} block. The former contains the reasoning process, whereas the latter contains the final answer. Whitespace is permitted before, between, and after the two blocks, but no other content is allowed. Let \(\mathcal{Y}_{\text{Fmt}}\) denote the set of outputs that satisfy this structure. The reward function is defined as:

{\small
\begin{equation}
R_{\text{Fmt}}(y) =
\begin{cases}
1, & y \in \mathcal{Y}_{\text{Fmt}}, \\
0, & \text{otherwise}.
\end{cases}
\end{equation}
}

\subsection{JSON Schema Validation}
The JSON Schema Validation reward is applied only to the payload extracted from the unique \texttt{<answer>...</answer>} block. Let $a(y)$ denote this payload. We parse $a(y)$ as a JSON object and validate it against the schema specified for the corresponding task. A reward of 1 is assigned if the payload is valid JSON and contains all required keys with the expected data types. A reward of 0.5 is assigned if it is valid JSON but fails to fully satisfy the schema because one or more required keys are missing or have incorrect data types. If no unique answer payload can be extracted, or if the extracted payload is not valid JSON, the reward is 0. The reward function is defined as:

{\small
\begin{equation}
R_{\text{JSON}}(y) =
\begin{cases}
1,   & \text{valid and schema-complete } a(y), \\
0.5, & \text{valid but schema-incomplete } a(y), \\
0,   & \text{missing or invalid } a(y).
\end{cases}
\end{equation}
}

\subsection{Statute Citation Format Verification}
Model outputs should include statute citations in a canonical format. If all statute citations are correctly formatted, the model receives a reward of 1. If the citations are missing or incorrect, the reward is 0. The reward function is defined as:

{\small
\begin{equation}
R_{\text{CFmt}}(y) =
\begin{cases}
1, & \text{correct citation format}, \\
0, & \text{incorrect format or missing}.
\end{cases}
\end{equation}
}

\subsection{Language Style Validation}
Model outputs are expected to adhere to a formal, neutral, and objective writing style. Because these stylistic attributes are difficult to assess reliably using rule-based methods alone, we employ Gemini-3-Pro-Preview as an LLM-based evaluator. Given a predefined evaluation prompt $P_{\text{Style}}$ specifying the assessment criteria, the evaluator assigns each response $y$ a style reward:

{\small
\begin{equation}
R_{\text{Style}}(y)
=
E_{\text{Style}}(y; P_{\text{Style}})
\in \{0, 0.5, 1\},
\end{equation}
}

where $E_{\text{Style}}$ denotes the LLM-based style evaluator. The reward is assigned according to the following criteria:
\begin{itemize}
\item A score of 1 is assigned when the response is formal, neutral, and objective.
\item A score of 0.5 is assigned when the response is generally formal, neutral, and objective, with only minor subjective or emotional expressions.
\item A score of 0 is assigned when the response is informal or contains substantial subjective or emotional language.
\end{itemize}

\subsection{Statute Authenticity Verification}
We verify the authenticity of every statute citation instance in the reasoning trace included in output $y$, requiring each cited statute to correspond to an actual legal provision. This reward builds on the Non-Hallucinated Statute Rate introduced in prior work~\cite{hu-etal-2025-fine}.

\begin{enumerate}
\item For each statute citation instance, extract the generated statute using regular expressions or LLMs, denoted as
$L_{\text{gen}}=\{S_{\text{name}}, S_{\text{number}}, S_{\text{content}}\}$,
where $S_{\text{name}}$ is the name of the statute, $S_{\text{number}}$ is the statute number, and $S_{\text{content}}$ is the content of the statute.

\item Embed the extracted content together with all statute texts in the reference database using SimBERT~\cite{SimbertBaseChinese}.

\item Calculate the semantic similarity between the generated content and each real statute. Select the real statute with the highest similarity as the best match and denote it by
$L_{\text{best}}=\{S'_{\text{name}}, S'_{\text{number}}, S'_{\text{content}}\}$.
The rationale behind this process is that although large models may not generate entirely hallucination-free statutes, they generally maintain high semantic consistency with real statutes. The closer the model-generated statute is to a real statute, the higher its semantic similarity.

\item Using rule-based comparisons, we determine whether the generated statute $L_{\text{gen}}$ is consistent with the matched real statute $L_{\text{best}}$. Specifically, $L_{\text{gen}}$ is considered non-hallucinated if the generated statute content occurs as a substring of the matched official statute text,
$S_{\text{number}}=S'_{\text{number}}$, and
$S_{\text{name}}\in\mathcal{A}(S'_{\text{name}})$,
where $\mathcal{A}(S'_{\text{name}})$ denotes the predefined set of accepted names and aliases for the matched statute.
\end{enumerate}

Let $N_{\text{cite}}(y)$ denote the total number of statute citation instances in the reasoning trace of output $y$, and let $N_{\text{nh}}(y)$ denote the number of non-hallucinated statute citation instances among them. We use the proportion of non-hallucinated statute citation instances as the reward score:

{\small
\begin{equation}
R_{\text{StatReal}}(y)
=
\frac{N_{\text{nh}}(y)}
{N_{\text{cite}}(y)}.
\end{equation}
}

If no statute citation instance is extracted, i.e., $N_{\text{cite}}(y)=0$, we set $R_{\text{StatReal}}(y)=0$. Otherwise, $R_{\text{StatReal}}(y)$ takes a value between 0 and 1. A higher value indicates that a larger proportion of the generated statute citation instances correspond to authentic legal provisions.

\subsection{Statute-Conclusion Consistency Verification}
We evaluate whether the statutes cited in each model response support its final conclusion. Because this relationship is difficult to assess reliably using rule-based methods alone, we employ Gemini-3-Pro-Preview as an LLM-based evaluator. Given a predefined evaluation prompt specifying the assessment criteria, the evaluator assigns each response a consistency reward according to the relationship between the cited statutes and the final conclusion:

{\small
\begin{equation}
R_{\text{Consistency}}(y) =
\begin{cases}
1,   & \text{direct support}, \\
0.5, & \text{potential support}, \\
0,   & \text{contradiction or irrelevance}.
\end{cases}
\end{equation}
}

A reward of 1 is assigned when the cited statutes directly support the conclusion, 0.5 when they provide potential support, and 0 when they contradict or are irrelevant to the conclusion.

\section{Identifying Reward-Hacking Behavior}
\label{sec:reward_hacking_diagnostic}

To examine whether continuously optimizing against a fixed auxiliary reward can induce reward-hacking behavior, we conduct a controlled diagnostic experiment.

\paragraph{Experimental Setup.}
We initialize both reinforcement-learning variants from the same JurisCoT-SFT checkpoint, which is also directly evaluated as the SFT baseline. During reinforcement learning, we retain the complete set of primary rewards used in the main experiments and introduce \(S_5\), Statute Authenticity Verification, as the only auxiliary reward.

We evaluate all three models on the same fixed subset of ChJGBench and report four metrics. In addition to the \textit{Law} and \(S_5\) reward scores, we report Avg. Statute Citations per Response and Avg. Mentions per Distinct Correct and Relevant Statute to determine whether an increase in the \(S_5\) reward score is driven by redundant citations or genuine improvements in legal-task performance.

The two reinforcement-learning variants use the same training data, sampling parameters, and primary rewards; their key distinction is how \(S_5\) is scheduled. In conventional GRPO, \(S_5\) remains active throughout training, whereas LABPO manages its lifecycle and retires it when its marginal contribution diminishes.

We compare three settings: (1) direct evaluation of JurisCoT-SFT without reinforcement learning; (2) JurisCoT-SFT followed by conventional GRPO using the primary rewards and \(S_5\); and (3) JurisCoT-SFT followed by LABPO using the same rewards.

\begin{table*}[t]
\centering
\small
\setlength{\tabcolsep}{4pt}
\renewcommand{\arraystretch}{1.15}
\begin{tabularx}{\textwidth}{lcc>{\centering\arraybackslash}X>{\centering\arraybackslash}X}
\toprule
\textbf{Model} & \textbf{\textit{Law}} & \makecell{\textbf{\(S_5\) Reward}\\\textbf{Score}} & \makecell{\textbf{Avg. Statute Citations}\\\textbf{per Response}} & \makecell{\textbf{Avg. Mentions per Distinct}\\\textbf{Correct and Relevant Statute}} \\
\midrule
JurisCoT-SFT & 0.667 & 0.542 & 3.18 & 1.21 \\
JurisCoT-SFT + GRPO & 0.669 & 0.934 & 12.46 & 4.87 \\
JurisCoT-SFT + LABPO & 0.681 & 0.761 & 4.27 & 1.48 \\
\bottomrule
\end{tabularx}
\caption{Results of the controlled diagnostic experiment on reward hacking using a fixed subset of ChJGBench.}
\label{tab:reward_hacking_diagnostic}
\end{table*}

\paragraph{Results.}
As shown in Table~\ref{tab:reward_hacking_diagnostic}, continuously optimizing against \(S_5\) under conventional GRPO increases the \(S_5\) reward from 0.542 to 0.934, while the task-level \textit{Law} score changes only marginally, from 0.667 to 0.669. Meanwhile, the average number of statute citations per response increases from 3.18 to 12.46, and the average number of mentions per distinct correct and relevant statute increases from 1.21 to 4.87.

This pattern reveals a clear reward-hacking strategy. Because $S_5$ is computed over individual statute citation instances, repeatedly citing an authentic statute adds to both the number of non-hallucinated citation instances and the total number of citation instances while leaving the number of hallucinated citation instances unchanged. Consequently, the model can increase the ratio-based $S_5$ reward by diluting the contribution of hallucinated citations, without meaningfully improving its performance on statutory provision application.

In comparison, LABPO substantially limits this behavior: the average number of statute citations per response decreases to 4.27, and the average number of mentions per distinct correct and relevant statute decreases to 1.48, while the \textit{Law} score increases to 0.681. Although its \(S_5\) reward is lower than that achieved by conventional GRPO, LABPO performs better on the underlying legal task.

\section{Reasoning Performance Comparison Details}  
\label{sec:ReasoningPerformanceComparisonDetails}

\begin{enumerate}
\item We use the test set described in Section~\ref{sec:Experimental_Setup} to generate complete reasoning traces for three different models: Qwen3-8B, Qwen3-8B with JurisCoT-SFT + LABPO, and DeepSeek-R1, all under the same prompt template. The generation settings are as follows: temperature = 0.7, top\_p = 0.95, and max\_new\_tokens = 8192.

\item For each comparison, we provide Gemini-3-Pro-Preview with an input formatted as \{query, ground-truth answer, reference reasoning trace, Model A reasoning trace, Model B reasoning trace\}. The evaluator is asked to produce a win, tie, or loss judgment for Model A relative to Model B along two dimensions: Helpfulness and Factual Correctness. To reduce sampling variability, we set the evaluator temperature to 0.

\item The definitions for these two evaluation dimensions are as follows: Helpfulness primarily evaluates whether the reasoning process is clear, logical, and contributes to solving the problem. Factual Correctness assesses whether the facts in the reasoning process are accurate and error-free. For Factual Correctness, Gemini-3-Pro-Preview uses the ground-truth answer and reference reasoning trace to determine which model's reasoning trace better aligns with the reference information or contains fewer factual discrepancies.

\item For any pairwise model comparison, the win rate for a specific dimension is defined as:

{\small
\begin{equation}
\text{Win Rate}
=
\frac{\text{Number of wins in that dimension}}
{\text{Total number of samples}}.
\end{equation}
}
\end{enumerate}

\section{Ablation Study of Signal-Rebalancing Objectives for SFT}
\label{app:sft_signal_rebalancing}

\paragraph{Experimental Setup.}
To better understand the source of the performance gains from JurisCoT-SFT at the SFT stage, we compare standard SFT, a signal-rebalancing baseline based on Answer-Token Upsampling, and JurisCoT-SFT under different values of $\lambda$. All experiments in this section exclude LABPO, allowing us to isolate the effects of the SFT objective. Across all variants, we use the same base model (Qwen3-8B), training data, batch size, learning-rate schedule, maximum sequence length, and number of training steps.

\paragraph{Standard SFT.}
For each training instance, let $x$ denote the input query, $t=(t_1,\ldots,t_{|t|})$ the reasoning trace, and $a=(a_1,\ldots,a_{|a|})$ the final answer. We define the unnormalized negative log-likelihood losses over the reasoning and answer tokens as follows:

{\small
\begin{equation}
\mathcal{L}_{t}
=
\sum_{i=1}^{|t|}
-\log p_\theta(t_i \mid x, t_{<i}).
\end{equation}
}

{\small
\begin{equation}
\mathcal{L}_{a}
=
\sum_{j=1}^{|a|}
-\log p_\theta(a_j \mid x, t, a_{<j}).
\end{equation}
}

The standard SFT objective minimizes the average token-level negative log-likelihood over the concatenated reasoning and answer sequence:

{\small
\begin{equation}
\mathcal{L}_{\mathrm{SFT}}
=
\frac{\mathcal{L}_{t} + \mathcal{L}_{a}}{|t| + |a|}.
\end{equation}
}

\paragraph{Answer-Token Upsampling.}
Because legal reasoning traces are often substantially longer than final answers, reasoning tokens may dominate the standard SFT objective, thereby attenuating the supervision provided by the final answer. We therefore consider an Answer-Token Upsampling baseline that increases the contribution of answer tokens by a fixed factor $k$:

{\small
\begin{equation}
\mathcal{L}_{\mathrm{AnsUp}}
=
\frac{\mathcal{L}_{t} + k\mathcal{L}_{a}}{|t| + k|a|}.
\end{equation}
}

We set $k=4$ because, on average, the reasoning traces in our data contain approximately four times as many tokens as the final answers. This baseline provides a simple heuristic for compensating for the attenuated answer-token signal. However, it requires $k$ to be selected a priori and cannot adapt to instance-specific variation in the relative lengths of the two segments.

\paragraph{JurisCoT-SFT.}
JurisCoT-SFT instead normalizes the reasoning and answer losses separately and combines them using a weighting coefficient $\lambda$:

{\small
\begin{equation}
\mathcal{L}_{\mathrm{JurisCoT-SFT}}
=
\lambda
\frac{\mathcal{L}_{t}}{|t|}
+
(1-\lambda)
\frac{\mathcal{L}_{a}}{|a|}.
\end{equation}
}

We evaluate $\lambda \in \{0.25,0.50,0.75\}$. The setting $\lambda=0.50$ assigns equal aggregate weight to the reasoning trace and the final answer. Smaller values place greater emphasis on the final answer, whereas larger values place greater emphasis on the reasoning trace.

\begin{table}[t]
\centering
\small
\begin{tabular}{lcc}
\toprule
\textbf{Objective} & \textbf{Setting} & \textbf{Avg.} $\uparrow$ \\
\midrule
Standard SFT & -- & 0.7399 \\
Answer-Token Upsampling & $k=4$ & 0.7428 \\
\midrule
JurisCoT-SFT & $\lambda=0.25$ & 0.7629 \\
JurisCoT-SFT & $\lambda=0.50$ & 0.7667 \\
JurisCoT-SFT & $\lambda=0.75$ & 0.7612 \\
\bottomrule
\end{tabular}
\caption{Ablation study of SFT objectives without LABPO. All variants are trained with Qwen3-8B under the same experimental configuration. Avg.\ denotes the average score on ChJGBench.}
\label{tab:sft_objective_ablation}
\end{table}

\paragraph{Results.}
Standard SFT achieves an average score of 0.7399. Answer-Token Upsampling with $k=4$ increases the score to 0.7428, an absolute improvement of 0.0029. This modest gain indicates that simply increasing the global weight of answer tokens provides only limited benefit.

By contrast, all three JurisCoT-SFT variants outperform both standard SFT and Answer-Token Upsampling. The best performance is achieved with $\lambda=0.50$, yielding a score of 0.7667 and absolute improvements of 0.0268 over standard SFT and 0.0239 over Answer-Token Upsampling. These results suggest that separately normalizing the two segment-level losses and explicitly controlling their relative contributions is more effective in this setting than either standard token-level averaging or fixed answer-token upsampling.

One possible explanation for why $\lambda=0.50$ performs best is the quality of the reasoning supervision. In our main experiments, reasoning traces are retained for training only after undergoing quality-control procedures such as Attempt-Bounded Self-Verification. Consequently, these traces provide relatively reliable supervision, making it reasonable to assign comparable aggregate weights to the reasoning and answer losses in the present setting. However, this result should not be interpreted as evidence that equal weighting is universally optimal. When reasoning traces are less reliable, smaller values of $\lambda$ may be preferable because they reduce the influence of noisy intermediate reasoning while placing greater emphasis on the final answer. This ability to adjust the balance according to the quality of the reasoning supervision illustrates the flexibility of the parameterized JurisCoT-SFT objective.

\section{Human Evaluation and Inter-Annotator Agreement}  
\label{sec:human_evaluation}

\paragraph{Human Evaluation Setup.}
We conduct a human evaluation to assess the reliability of the proposed reasoning evaluation protocol and to analyze inter-annotator consistency. Three annotators with formal legal training independently evaluate 500 queries sampled from the test set using a stratified sampling strategy. To ensure representativeness, the queries are stratified by (i) legal domain, including criminal, civil, and administrative cases, and (ii) case complexity, measured by the length of the factual description.
Each query corresponds to one pairwise comparison instance. The 500 instances are evenly divided into two comparison settings: our proposed method versus DeepSeek-R1 and our proposed method versus Qwen3-8B, ensuring a balanced number of pairs for both baselines.

\paragraph{Annotation Protocol.}
For each query, annotators perform pairwise comparisons between two anonymized reasoning traces along two dimensions: \textbf{Helpfulness} and \textbf{Factual Correctness}. The order of the two candidates is randomized for each instance to avoid positional bias. All annotations are conducted with annotators blinded to the model identities associated with the reasoning traces.

\paragraph{Aggregation Rule.}

For each query and each evaluation dimension, annotators assign one of three labels: \(\{A\ \text{wins},\ \text{Tie},\ B\ \text{wins}\}\). Final human labels are determined using a conservative majority rule: a model is assigned a \emph{win} label only if at least two out of three annotators explicitly prefer its reasoning trace. If neither model receives at least two win votes, the final label is determined as \emph{Tie}. This aggregation strategy reduces the impact of subjective or ambiguous judgments and yields more reliable human evaluation results.

\paragraph{Agreement Metrics.}

We report a set of standard metrics to evaluate both human annotator consistency and human--LLM alignment.

For human inter-annotator agreement, we use Krippendorff's $\alpha$ and Majority Agreement. Krippendorff's $\alpha$ measures chance-corrected agreement among multiple annotators and is suitable for nominal labels. In our setting, the three comparison outcomes (\emph{A wins}, \emph{Tie}, \emph{B wins}) are treated as nominal categories. Majority agreement measures the percentage of queries for which at least two of the three annotators assign the same label, indicating whether a clear dominant human preference exists and supporting the use of majority voting for label aggregation.

To assess the reliability of the LLM evaluator (Gemini-3-Pro-Preview), we compare its judgments with the aggregated human labels using Accuracy and Cohen's $\kappa$. Accuracy measures the proportion of exact matches between the LLM prediction and the human consensus, while Cohen's $\kappa$ provides a chance-corrected estimate of agreement between the two.

\begin{table}[t]
\centering
\small
\renewcommand{\arraystretch}{0.9}
\begin{tabular}{cccc}
\toprule
$\boldsymbol{\alpha}$ &
\textbf{Majority} &
\textbf{LLM Acc.} &
\textbf{LLM $\kappa$} \\
\midrule
0.615 &
96.0\% &
0.84 &
0.756 \\
\bottomrule
\end{tabular}
\caption{Human inter-annotator agreement and human--LLM alignment results on 500 queries.}
\label{tab:agreement}
\end{table}

\paragraph{Results.}
Table~\ref{tab:agreement} summarizes human inter-annotator agreement and human--LLM alignment. Krippendorff's $\alpha=0.615$ indicates a reasonably good level of agreement among the annotators, while the 96.0\% majority agreement shows that at least two annotators selected the same label for most queries. The LLM evaluator achieves an accuracy of 0.84 against the aggregated human labels, with Cohen's $\kappa=0.756$ indicating substantial agreement under the same evaluation protocol. These results support the use of the LLM evaluator as an automatic judge for this task.

\section{Ablation Studies on LawBench and LegalBench}
\label{sec:Ablation_Studies_on_LawBench_and_LegalBench}

Table~\ref{tab:ablation_study_lawbench} and Table~\ref{tab:ablation_study_legalbench} report the ablation results on LawBench and LegalBench, respectively, using Qwen3-8B as the backbone model.

\begin{table}[t]
\centering
\small
\setlength{\tabcolsep}{3pt}
\renewcommand{\arraystretch}{1}
\begin{tabular}{c cc ccc r}
\toprule
\multirow{2}{*}{\textbf{No.}} 
& \multicolumn{2}{c}{\textbf{SFT}} 
& \multicolumn{3}{c}{\textbf{RL}} 
& \multirow{2}{*}{\textbf{Avg.} $\uparrow$} \\
\cmidrule(lr){2-3}
\cmidrule(lr){4-6}
& Used & Variant & Used & Retire & Rollback & \\
\midrule
1 & \checkmark & \checkmark & \checkmark & \checkmark & \checkmark & 0.7893 \\
2 & \checkmark & \checkmark & $\times$ & $\times$ & $\times$ & 0.7593 \\
3 & \checkmark & $\times$ & \checkmark & \checkmark & \checkmark & 0.7651 \\
4 & \checkmark & $\times$ & $\times$ & $\times$ & $\times$ & 0.7484 \\
5 & $\times$ & $\times$ & $\times$ & $\times$ & $\times$ & 0.7220 \\
\bottomrule
\end{tabular}
\caption{Ablation study on LawBench. A \checkmark indicates that the component is enabled, while $\times$ indicates that it is disabled. Average denotes the mean of all metrics in the range [0, 1].}
\label{tab:ablation_study_lawbench}
\end{table}

\begin{table}[t]
\centering
\small
\setlength{\tabcolsep}{3pt}
\renewcommand{\arraystretch}{1}
\begin{tabular}{c cc ccc r r}
\toprule
\multirow{2}{*}{\textbf{No.}} 
& \multicolumn{2}{c}{\textbf{SFT}} 
& \multicolumn{3}{c}{\textbf{RL}} 
& \multirow{2}{*}{\textbf{Avg.} $\uparrow$} 
& \multirow{2}{*}{\textbf{Sara-N} $\downarrow$} \\
\cmidrule(lr){2-3}
\cmidrule(lr){4-6}
& Used & Var & Used & Ret & Roll & & \\
\midrule
1 & \checkmark & \checkmark & \checkmark & \checkmark & \checkmark & 0.826 & 1.753 \\
2 & \checkmark & \checkmark & $\times$ & $\times$ & $\times$ & 0.777 & 3.949 \\
3 & \checkmark & $\times$ & \checkmark & \checkmark & \checkmark & 0.787 & 5.906 \\
4 & \checkmark & $\times$ & $\times$ & $\times$ & $\times$ & 0.759 & 7.231 \\
5 & $\times$ & $\times$ & $\times$ & $\times$ & $\times$ & 0.735 & 8.093 \\
\bottomrule
\end{tabular}
\caption{Ablation study on LegalBench. A \checkmark indicates that the component is enabled, and $\times$ indicates that it is disabled. Abbreviations: Var = Variant, Ret = Retire, Roll = Rollback. Average denotes the mean of all accuracy metrics. Sara-N corresponds to the Log MSE metric and is therefore reported separately.}
\label{tab:ablation_study_legalbench}
\end{table}

\section{Training Dataset Statistics}
\label{sec:TrainingDatasetStatistics}

\subsection{Statutory Law Dataset}

Table~\ref{tab:statute_corpus_stats} provides detailed statistics of the statutory law training dataset.

\begin{table}[t]
  \centering
  \small
  \begin{tabular}{lr}
    \toprule
    \textbf{Item} & \textbf{Count} \\
    \midrule
    Training instances \((x,t,a)\) & 2{,}734{,}100 \\
    Distinct statutes (clause level) & 35{,}010 \\
    Causes of action & \(>400\) \\
    \midrule
    Criminal cases & 581{,}579 \\
    Civil cases & 1{,}625{,}584 \\
    Administrative cases & 526{,}937 \\
    \midrule
    Average token count per reasoning trace & 3{,}922.4 \\
    Average token count per answer & 902.2 \\
    \bottomrule
  \end{tabular}
  \caption{Statistics of the statutory-law training dataset, including the total number of training instances, the number of distinct statutes at the clause level, the distribution of legal cases, and the average token counts per reasoning trace and per answer.}
  \label{tab:statute_corpus_stats}
\end{table}

\subsection{Case Law Dataset}

Table~\ref{tab:case_corpus_stats} provides detailed statistics of the case law training dataset.

\begin{table}[t]
  \centering
  \small
  \begin{tabular}{lr}
    \toprule
    \textbf{Item} & \textbf{Count} \\
    \midrule
    Training instances \((x,t,a)\) & 35{,}965 \\
    \midrule
    Average token count per reasoning trace & 2{,}369.2 \\
    Average token count per answer & 3.5 \\
    \bottomrule
  \end{tabular}
  \caption{Statistics of the case-law training dataset, including the total number of training instances and the average token counts in reasoning traces and answers.}
  \label{tab:case_corpus_stats}
\end{table}

\section{Detailed Definitions of Evaluation Metrics}
\label{sec:EvaluationMetrics}

\subsection{ChJGBench Evaluation Metrics}

All diagnostic dimensions used in this work, together with their corresponding evaluation metrics, are adopted from ChJGBench~\cite{hu_2026_22171576}. We refer readers to the original benchmark for their detailed definitions.

To account for the unequal numbers of diagnostic dimensions across litigation domains, we introduce \textbf{\textsc{Average}}, a domain-balanced hierarchical macro-averaging metric. Directly averaging all dimension-level scores would give greater weight to domains evaluated with more dimensions. We therefore first average the dimension-level scores within each domain and then assign equal weight to the resulting three domain-level scores. The resulting \textbf{\textsc{Average}} scores are reported in Tables~\ref{tab:main_results_cn} and~\ref{tab:ablation_study}.

Let \(M_{\mathrm{shared}}=\{\mathrm{Law},\mathrm{ISS},\mathrm{JUS}\}\), and let \(M_d\) denote the domain-specific metrics. For the criminal, civil, and administrative domains, these metrics are \(\{\mathrm{Fine},\mathrm{MIT},\mathrm{Charge},\mathrm{Pen\_M},\mathrm{REC}\}\), \(\{\mathrm{SUP},\mathrm{PAY}\}\), and \(\{\mathrm{DIS}\}\), respectively. Each domain-level score averages its shared and domain-specific dimensions:

{\small
\begin{equation}
S_d
=
\frac{1}{|M_{\mathrm{shared}} \cup M_d|}
\sum_{m \in M_{\mathrm{shared}} \cup M_d} s_m.
\end{equation}
}

The final \textsc{Average} score assigns equal weight to the three domains:

{\small
\begin{equation}
\textsc{Average}
=
\frac{
S_{\mathrm{Crim}} +
S_{\mathrm{Civ}} +
S_{\mathrm{Adm}}
}{3}.
\label{eq:weighted_avg_score}
\end{equation}
}

\subsection{LawBench Evaluation Metrics}

We select three representative Legal Judgment Prediction tasks from \textbf{LawBench} for evaluation. For classification-based tasks, namely Article Prediction and Charge Prediction, we use \textbf{F1} as the evaluation metric.

For Sentence Prediction, we use the official LawBench \textbf{Normalized Log-Distance (NLD)} metric~\cite{fei-etal-2024-lawbench}. Let $N_{\mathrm{eval}}$ be the number of fixed-term-imprisonment cases after excluding references involving the death penalty or life imprisonment, and let $y_i$ be the nonnegative reference term in months. After converting Chinese numerals, the evaluator parses the first duration in months, or the first duration in years multiplied by 12 if no month duration is found, as the nonnegative integer $\hat{y}_i$. It then computes

{\small
\begin{equation}
d_i =
\begin{cases}
\left|\log(\hat{y}_i+1)-\log(y_i+1)\right|,
& \hat{y}_i\ \text{is parsed},\\
\log 216,
& \text{otherwise},
\end{cases}
\end{equation}

\begin{equation}
\mathrm{NLD}
=
1-
\frac{1}{N_{\mathrm{eval}}\log 216}
\sum_{i=1}^{N_{\mathrm{eval}}} d_i.
\end{equation}
}

If no duration is parsed, the response is an abstention and contributes an instance-level score of zero.

\subsection{LegalBench Evaluation Metrics}

For classification-based tasks in \textbf{LegalBench}, we adopt the \textbf{Accuracy (Acc)} metric to measure prediction correctness. Accuracy is defined as the proportion of correctly predicted samples among all evaluated instances:

{\small
\begin{equation}
\mathrm{Acc}
=
\frac{1}{N}
\sum_{i=1}^{N}
\mathbb{I}\!\left(\hat{y}_i = y_i\right),
\end{equation}
}

where $N$ denotes the total number of samples, $\hat{y}_i$ is the predicted label, $y_i$ is the ground-truth label, and $\mathbb{I}(\cdot)$ is the indicator function that returns 1 when the prediction matches the ground truth and 0 otherwise.

For numerical reasoning tasks such as \textbf{Sara-N}, we employ \textbf{Log Mean Squared Error (Log MSE)} to evaluate prediction accuracy on continuous numerical outputs. Log MSE measures the squared difference between the logarithms of predicted and ground-truth values, which helps stabilize variance when values span multiple orders of magnitude:

{\small
\begin{equation}
\mathrm{LogMSE}
=
\frac{1}{N}
\sum_{i=1}^{N}
\left(
\log(\hat{y}_i) - \log(y_i)
\right)^2,
\end{equation}
}

where $N$ denotes the number of samples, $\hat{y}_i$ is the predicted numerical value, and $y_i$ is the corresponding ground-truth value.

\begin{table*}[t]
\centering
\small
\setlength{\tabcolsep}{8pt}
\begin{tabularx}{\textwidth}{@{}l >{\raggedright\arraybackslash}X c c c@{}}
\toprule
\textbf{Task} & \textbf{Task Full Name} & \textbf{Litigation Type} & \textbf{Benchmark} & \textbf{Metric} \\
\midrule

\multicolumn{5}{c}{\textit{\underline{Statutory Law Tasks}}} \\

Law       & Statutory Provision Application                & General        & ChJGBench & Case-level set F1 \\
ISS       & Disputed Issues Identification                 & General        & ChJGBench & LLM-as-a-judge \\
JUS       & Judicial Justification                         & General        & ChJGBench & LLM-as-a-judge \\

Charge    & Criminal Charge Prediction                     & Criminal       & ChJGBench & Defendant-level set F1 \\
Fine      & Monetary Fine                                  & Criminal       & ChJGBench & NVS  \\
REC       & Recidivism Assessment                          & Criminal       & ChJGBench & Binary macro-F1 \\
MIT       & Mitigating Circumstances Identification        & Criminal       & ChJGBench & Binary macro-F1 \\
Pen\_M    & Penalty Duration Prediction                    & Criminal       & ChJGBench & NVS \\
SUP       & Civil Claim Support Determination              & Civil          & ChJGBench & Three-class macro-F1 \\
PAY       & Payment Obligation                             & Civil          & ChJGBench & NVS  \\
DIS       & Administrative Disposition Prediction          & Administrative & ChJGBench & LLM-as-a-judge \\

\midrule
AP        & Article Prediction                             & Criminal       & LawBench   & F1  \\
CP        & Charge Prediction                              & Criminal       & LawBench   & F1  \\
SP        & Sentence Prediction                            & Criminal       & LawBench   & NLD \\

\midrule

\multicolumn{5}{c}{\textit{\underline{Case Law Tasks}}} \\

LRC       & Legal Reasoning Causality                      & - & LegalBench & Accuracy \\
NYSJE     & New York State Judicial Ethics                 & - & LegalBench & Accuracy \\
Sara-N    & SARA Numeric Reasoning                         & - & LegalBench & Log MSE \\

\bottomrule
\end{tabularx}

\caption{Summary of evaluation tasks, including statutory law tasks and case law tasks, with their corresponding litigation types, benchmark sources, and evaluation metrics. In ChJGBench~\cite{hu_2026_22171576}, NVS stands for Normalized Value Similarity.}
\label{tab:tasks_metrics}
\end{table*}

\section{Implementation Details}
\label{sec:implementation_details}

We adopt a two-stage training strategy. Training is conducted on 32 NVIDIA A100 GPUs with 80 GB of memory each.

In the first stage, we perform SFT on the Qwen3-8B and Llama-3.1-8B base models with a learning rate of \(1\times10^{-5}\) and a sequence length cutoff of 16,384. The per-device training batch size is set to 1, with 2 gradient accumulation steps. We use a cosine learning-rate scheduler with a warmup ratio of 0.1. The balancing coefficient is set to \(\lambda=0.5\).

In the second stage, we apply reinforcement learning with a learning rate of \(1\times10^{-6}\), a global training batch size of 256, and a micro batch size of 128. In all main experiments, we set the target number of retained optimization steps to \(N=3000\), the minimum gain threshold to \(\epsilon=0.0005\), the patience window to \(H=4\), and the evaluation interval to \(n=100\).

LABPO distinguishes retained steps from total executed steps. The retained-step counter \(s\) determines progress toward \(N\) and is restored with the complete lifecycle-best checkpoint, including the policy parameters, optimizer state, and learning-rate scheduler state. The total executed-step count is not restored and is used only for overhead accounting. Consequently, steps discarded by an auxiliary-signal rollback are re-executed during the subsequent training phase.

The primary reward remains active throughout reinforcement learning, while at most one auxiliary reward is active at any time. Development-set evaluation is performed after every \(n=100\) retained-step increments. Each phase maintains a running-best development score and its corresponding checkpoint. Once \(r\geq H\) and \(B_r-B_{r-H}<\epsilon\), the active auxiliary signal is retired, the lifecycle-best checkpoint is restored, and the next auxiliary signal is activated. The phase-local evaluation history is then reset. The stored score of the restored checkpoint initializes the new phase, avoiding an additional development-set evaluation.

After the last auxiliary signal has been retired, all auxiliary rewards are disabled and training continues using only the primary reward. The same monitoring criterion is applied during this final phase. If the criterion is satisfied, the phase-best checkpoint is restored and training terminates immediately without re-executing the discarded steps; otherwise, training continues until the retained-step counter reaches \(N\).

The development set is used strictly for model selection, including auxiliary-signal retirement, checkpoint rollback, and early-termination decisions. It is never used for gradient updates or for optimizing the policy or reward models. The training, development, and test sets are strictly disjoint, and all final results are evaluated on a completely held-out test set that is never accessed during training or model selection.

\section{Computational Overhead Analysis}
\label{sec:ComputationalOverheadAnalysis}

Compared with standard GRPO, LABPO introduces additional overhead from re-executing optimization steps discarded by auxiliary-signal rollback, periodic development-set evaluation, and checkpoint I/O.

Let \(N\) denote the target number of retained optimization steps, \(U\) the total number of executed optimization steps, \(t_{\mathrm{step}}\) the average time per optimization step, and \(t_{\mathrm{eval}}\) the time per development-set evaluation. For auxiliary-signal rollback \(a\), let \(r_a\) and \(r_a^\star\) denote the retirement index and the lifecycle-best checkpoint index, respectively. Since retirement is checked after every evaluation, the restored checkpoint can be at most \(H\) evaluation intervals before the retirement point. Therefore,

{\small
\begin{equation}
\begin{aligned}
d_a
&= n(r_a - r_a^\star) \leq Hn, \\
D_{\mathrm{aux}}
&= \sum_{a=1}^{A_{\mathrm{aux}}} d_a
\leq A_{\mathrm{aux}} Hn
\leq mHn,
\end{aligned}
\end{equation}
}

where \(A_{\mathrm{aux}}\leq m\) is the number of auxiliary-signal rollbacks and \(D_{\mathrm{aux}}\) is the total number of re-executed steps. A rollback during the final phase in which only the primary reward is used is not included in \(D_{\mathrm{aux}}\), because it is followed immediately by termination.

If training reaches \(N\) retained steps, then \(U=N+D_{\mathrm{aux}}\). Early termination during the final phase can only reduce \(U\). The development set is evaluated once before training and whenever the retained-step counter reaches a multiple of \(n\), including repeated evaluations when discarded intervals are re-executed. Because the stored score of a restored checkpoint is reused at phase transitions,

{\small
\begin{equation}
\begin{aligned}
U
&\leq N + D_{\mathrm{aux}}
\leq N + mHn, \\
E(U)
&= 1 + \left\lfloor \frac{U}{n} \right\rfloor
\leq
1 + \left\lfloor \frac{N + mHn}{n} \right\rfloor,
\end{aligned}
\end{equation}
}

where \(E(U)\) is the total number of development-set evaluations.

The computational costs of standard GRPO and LABPO are

{\small
\begin{equation}
\begin{aligned}
C_{\mathrm{GRPO}}
&= N t_{\mathrm{step}}, \\
C_{\mathrm{LABPO}}
&= U t_{\mathrm{step}}
+ E(U) t_{\mathrm{eval}}
+ C_{\mathrm{io}},
\end{aligned}
\end{equation}
}

where \(C_{\mathrm{io}}\) denotes the scheduling-related cost of checkpoint saving, loading, and restoration. LABPO performs at most \(m\) auxiliary-signal restorations and one additional restoration during the final phase in which only the primary reward is used.

Under our configuration of \(N=3000\), \(m=6\), \(H=4\), and \(n=100\), the worst-case bounds are

{\small
\begin{equation}
\begin{aligned}
D_{\mathrm{aux}} &\leq 2400, \\
U &\leq 5400, \\
E(U) &\leq 55, \\
C_{\mathrm{LABPO}}
&\leq
5400 t_{\mathrm{step}}
+ 55 t_{\mathrm{eval}}
+ C_{\mathrm{io}}.
\end{aligned}
\end{equation}
}

The corresponding relative overhead over standard GRPO is bounded by

{\small
\begin{equation}
\begin{aligned}
\mathrm{Overhead}
&=
\frac{C_{\mathrm{LABPO}} - C_{\mathrm{GRPO}}}
{C_{\mathrm{GRPO}}} \\
&\leq
0.8
+
\frac{55}{3000}
\frac{t_{\mathrm{eval}}}{t_{\mathrm{step}}}
+
\frac{C_{\mathrm{io}}}
{3000 t_{\mathrm{step}}}.
\end{aligned}
\end{equation}
}

Thus, auxiliary-signal rollback increases the optimization-step computation by at most \(80\%\), corresponding to at most \(1.8\times\) the optimization-step cost of standard GRPO before evaluation and checkpoint I/O are included. This is a theoretical worst-case bound; the actual re-execution overhead \(D_{\mathrm{aux}}/N\) is smaller when the lifecycle-best checkpoint is close to the retirement point.

\section{Case Study}

\begin{figure*}[t]
    \centering
    \includegraphics[width=\textwidth, trim=5 5 5 0, clip]{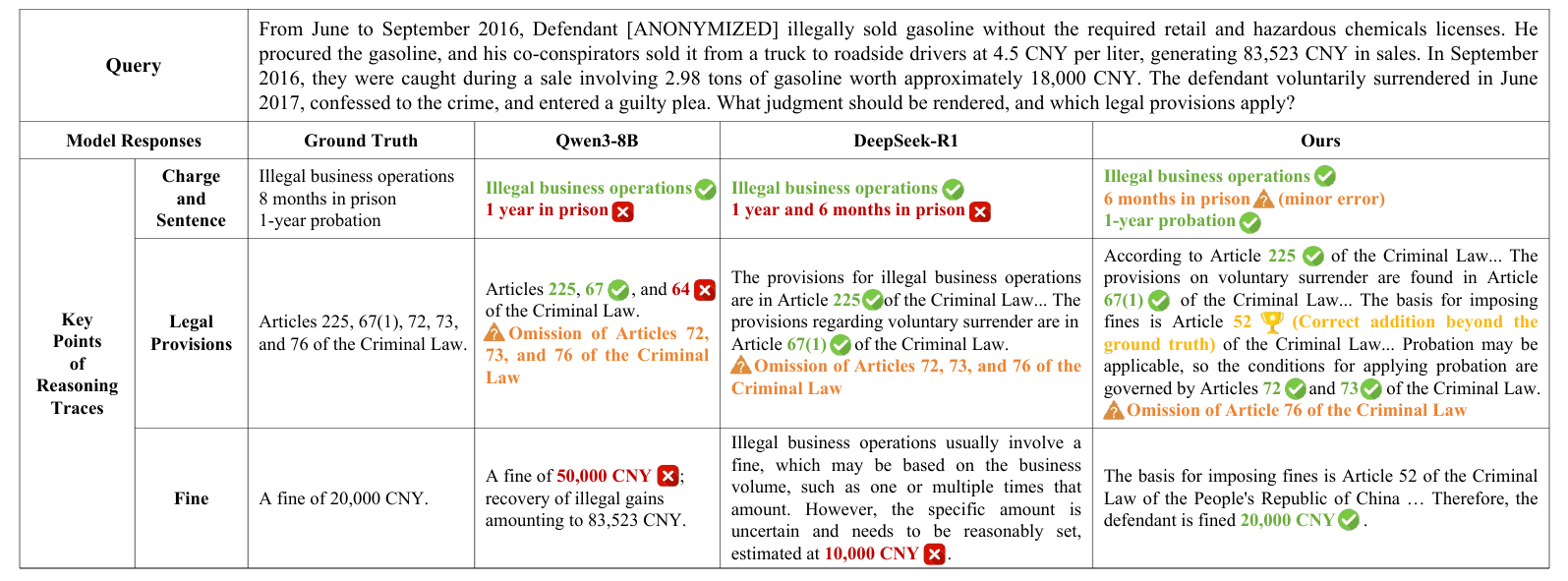}
    \caption{Case study: legal reasoning accuracy and completeness in a judgment concerning illegal business operations.}
    \label{fig:CaseStudy}
\end{figure*}

To assess the model's accuracy in both reasoning traces and final answers on complex legal reasoning tasks, we examine the representative case shown in Figure~\ref{fig:CaseStudy}. This example is drawn from a typical criminal case involving illegal business operations and comprises multiple subtasks, including charge identification, sentencing determination, application of legal provisions, and fine calculation. We compare the outputs produced by the model trained using our proposed algorithm with those generated by two open-source models, Qwen3-8B and DeepSeek-R1, across three dimensions: Charge and Sentence, Legal Provisions, and Fine.

The case shows that our model correctly identifies the defendant's voluntary surrender and eligibility for probation, with only a minor two-month discrepancy between the predicted prison sentence (six months) and the reference sentence (eight months). However, it omits one legal provision governing probation supervision (Article 76 of the Criminal Law). In contrast, both Qwen3-8B and DeepSeek-R1 fail to recognize the possibility of probation and instead predict prison sentences of 12 and 18 months, respectively. Additionally, both models omit legal provisions related to probation, including Articles 72, 73, and 76. This incomplete coverage of applicable legal provisions is reflected in incomplete sentencing rationales: both models predict imprisonment without considering probation. This example illustrates the importance of comprehensively recalling applicable legal provisions for reasonable sentencing.

Regarding fine determination, Qwen3-8B predicts a fine of 50,000 CNY, and DeepSeek-R1 predicts a fine of 10,000 CNY; both differ substantially from the reference fine of 20,000 CNY. In contrast, our model matches the reference amount. Notably, before generating the final answer, our model correctly identifies and applies Article 52 of the Criminal Law, providing a more comprehensive legal basis for fine determination than the reference answer. This suggests that reliable numerical decision-making in legal tasks depends not only on strong multi-step reasoning but also on the integration of domain-specific legal knowledge or structural constraints.

In summary, this example illustrates how general-purpose LLMs may exhibit inadequate sensitivity to procedural and conditional legal rules and weak numerical reasoning in legal contexts. In this case, the proposed training method mitigates these issues, improving accuracy across key judgment elements while maintaining interpretable reasoning traces.

\end{document}